\documentclass[lettersize,journal]{IEEEtran}
\usepackage{amsmath,amsfonts}
\usepackage{algorithmic}
\usepackage{algorithm}
\usepackage{array}
\usepackage[caption=false,font=normalsize,labelfont=sf,textfont=sf]{subfig}
\usepackage{textcomp}
\usepackage{stfloats}
\usepackage{url}
\usepackage{verbatim}
\usepackage{graphicx}
\usepackage{cite}
\usepackage{booktabs}
\usepackage{multirow}
\usepackage{bm}
\usepackage{capt-of}
\newcommand\blfootnote[1]{%
\begingroup 
\renewcommand\thefootnote{}\footnote{#1}%
\addtocounter{footnote}{-1}%
\endgroup 
}

\usepackage[table,xcdraw,dvipsnames]{xcolor}
\definecolor{TableDarkGreen}{RGB}{182,215,168}
\definecolor{TableLightGreen}{RGB}{207,234,215}
\newcommand{\red}[1]{$_{\color{RedOrange}\uparrow #1}$}
\begin{document}

\title{Continual Hand-Eye Calibration for Open-world Robotic Manipulation}

\author{Fazeng~Li, 
        Gan~Sun*, 
        Chenxi~Liu, 
        Yao~He,
        Wei~Cong,
        and~Yang~Cong
}


\twocolumn[{%
\renewcommand\twocolumn[1][]{#1}%
\maketitle
\includegraphics[width=1\linewidth]{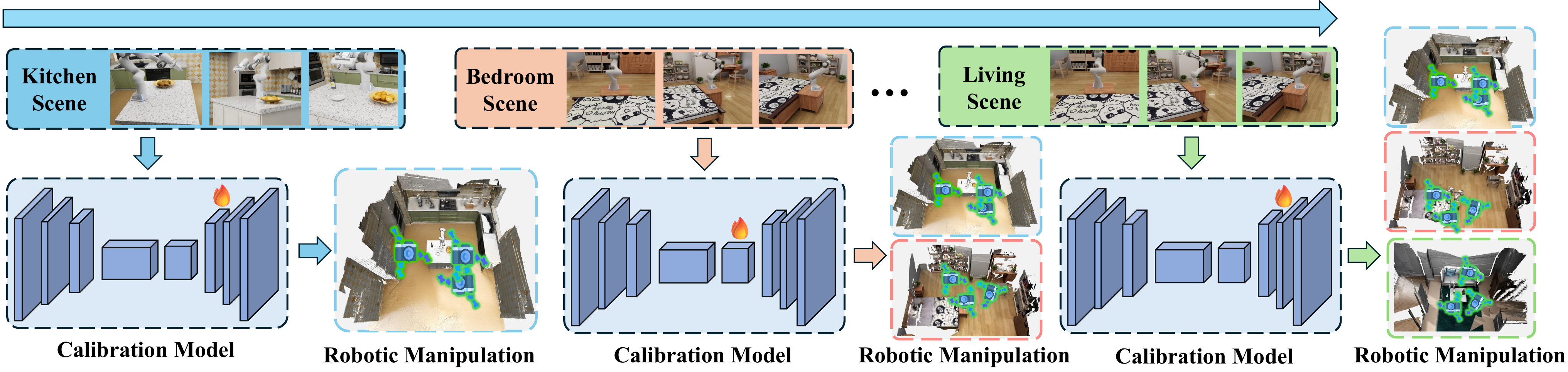}
\captionof{figure}{Motivation of continual hand-eye calibration problem, where a single calibration model is sequentially trained on several open-world scenes (\emph{e.g.}, Kitchen, Bedroom, $\cdots$, Living Room). The attempt of this work is to maintain accurate camera pose estimation across all previously encountered scenes while adapting to new scene, supporting persistent robotic manipulation without recalibration. \vspace{1em}}
\label{fig:teaser}
}]
\blfootnote{This work is supported by the National Key Research and Development Program of China (2022YFC2806101) and National Nature Science Foundation of China under Grant (62273333, 62225310, 62127807), and the Fundamental Research Funds for the Central Universities (2024ZYGXZR024).}
\blfootnote{Fazeng Li, Gan Sun, Yao He and Yang Cong are with the School of Automation Science and Engineering, South China University of Technology, Guangzhou 510641, China. (email: lifazeng818@gmail.com, sungan1412@gmail.com, heyao0293@gmail.com, congyang81@gmail.com)}
\blfootnote{Chenxi Liu and Wei Cong are with the State Key Laboratory of Robotics, the Institutes for Robotics and Intelligent Manufacturing, Chinese Academy of Sciences, Shenyang 110169, China, and also with the University of Chinese Academy of Sciences, Beijing 100049, China. (email: liuchenxi0101@gmail.com, congwei@sia.cn)}
\blfootnote{$^{*}$The corresponding author is \emph{Prof. Gan Sun}.}

\begin{abstract}
Hand-eye calibration through visual localization is a critical capability for robotic manipulation in open-world environments.
However, most deep learning-based calibration models suffer from catastrophic forgetting when adapting into unseen data amongst open-world scene changes, while simple rehearsal-based continual learning strategy cannot well mitigate this issue.
To overcome this challenge, we propose a continual hand-eye calibration framework, enabling robots to adapt to sequentially encountered open-world manipulation scenes through spatially replay strategy and structure-preserving distillation. 
Specifically, a Spatial-Aware Replay Strategy (SARS) constructs a geometrically uniform replay buffer that ensures comprehensive coverage of each scene pose space, replacing redundant adjacent frames with maximally informative viewpoints.
Meanwhile, a Structure-Preserving Dual Distillation (SPDD) is proposed to decompose localization knowledge into coarse scene layout and fine pose precision, and distills them separately to alleviate both types of forgetting during continual adaptation.
As a new manipulation scene arrives, SARS provides geometrically representative replay samples from all prior scenes, and SPDD applies structured distillation on these samples to retain previously learned knowledge. After training on the new scene, SARS incorporates selected samples from the new scene into the replay buffer for future rehearsal, allowing the model to continuously accumulate multi-scene calibration capability.
Experiments on multiple public datasets show significant anti scene forgetting performance, maintaining accuracy on past scenes while preserving adaptation to new scenes, confirming the effectiveness of the framework.
\end{abstract}


\section{Introduction}
\IEEEPARstart{V}{isual} localization-based hand-eye calibration estimates the spatial transformation between a camera and a robot by regressing camera poses from images, enabling precise mapping from visual observations to physical actions~\cite{tian2018active,li2024auto,jin2025nonsingular}. This capability is indispensable across a broad spectrum of robotic manipulation tasks. For instance, hand-eye calibration in industrial assembly task allows robots to align sub-millimeter components on production lines under varying lighting and tooling configurations~\cite{luo2024calibrator,stefanov2026perception,ma2026large}; reliable camera-to-robot pose estimation for autonomous grasping could enable robust object picking in cluttered and unstructured environments (\emph{e.g.,} warehouse logistics and bin sorting~\cite{raj2023scalable,ai2025review,huang2025you}); flexible manufacturing demands that hand-eye calibration generalize across diverse scenes where product types and workspace layouts change frequently, maintaining manipulation accuracy without manual recalibration~\cite{yang2023digital,liu2025aligning}. Amongst all these above scenarios, calibration accuracy directly determines the success rate and precision of robotic manipulation.

However, recent scene coordinate regression (SCR) methods for hand-eye calibration (\emph{e.g.,} ACE~\cite{ace} and GLACE~\cite{wang2024glace}) are designed for fixed environments.  
They implicitly encode scene geometry into neural networks and achieve accurate camera pose estimation within individual scenes, which is hard to fit the new environments from open-world scenes flow without retraining.
As shown in Figure~\ref{fig:teaser}, the scenes in open-world robotic applications could change continuously, most recently-proposed methods will fail to adapt to new scenes without forgetting previously learned ones.
\textit{For example, the model trained on kitchen scenes will get calibration result quickly even the position of the camera changed.  
However, when this service robot moves to a bedroom, the model will adapt to the new scene but forget the previous kitchen scene, which causes the robot to lose its calibration accuracy when it comes back to kitchen scene.}
This single-scene limitation fundamentally restricts the deployment of visual localization-based calibration in open-world robotic applications where environments change continuously.
Continual machine learning~\cite{wang2024comprehensive,li2024continual} is a usually straightforward solution to the forgetting issue in visual localization, as it allows models to adapt to new scenes while retaining performance on previously learned scenes.
For example, Wang et al.~\cite{wang2021continual} introduce Buff-CS, a method that constructs a replay buffer targeting scene coverage by selecting images based on subscene divisions. However, the subscene division relies heavily on the 3D pseudo-labels constructed by HSCNet\cite{li2020hierarchical}, resulting in computationally costly and difficult to transfer to more efficient SCR methods; Bahadir et al.~\cite{bahadir2024continual} formulate the problem of continual hand-eye calibration by partitioning a single scene into multiple robot-centered regions and treating each region as a separate continual task. Although it is effective in handling viewpoint changes within a fixed environment, this setting remains gap from open-world manipulation.

Inspired by this scenario, this paper aims to construct a continual localization-based hand-eye calibration framework that enables a single model to sequentially learn new scenes without forgetting previously learned environments.
To achieve this goal, we consider the challenges of continual hand-eye calibration framework from two perspectives: 
\begin{itemize}
  \item \textbf{Open-world Scene Forgetting}: the calibration model tends to forget previously learned scenes after training on a new scene. \textit{For instance, after being deployed in the Bedroom scene, the model may forget the Kitchen scene learned earlier, which fails to meet the requirements of open-world applications.} Reservoir sampling facilitates the construction of a replay buffer for revisiting previously seen scenes, but it may neglect unsampled regions. Therefore, how to construct a replay buffer that captures rich structural information is an important challenge.
  \item \textbf{Pose Precision Degradation}: the model may retain coarse scene awareness while losing fine-grained pose precision, leading to large camera pose errors. \textit{For example, the model may correctly recognize the Kitchen scene, yet estimate the camera pose in the coordinate system of the Bedroom scene, resulting in severe pose confusion.} Directly distilling the final outputs of the teacher model is a common strategy. Therefore, how to design an effective distillation pipeline remains a significant challenge.
\end{itemize}

To address these challenges, we propose a continual hand-eye calibration framework by comprising spatial-based replay buffer construction and structure-preserving knowledge distillation, which can achieve the robots manipulation in continuous open-world scenes. 
To achieve the goal of anti-forgetting in visual localization, we observe that the key is to maximize the geometric coverage of replay samples and separately preserve coarse pose knowledge and fine pose details. Specifically, a Spatial-Aware Replay Strategy~(SARS) is proposed to construct a geometrically diverse replay buffer via Poisson disk sampling in the joint position-orientation space, which ensures comprehensive coverage of each scene pose space, and further allivetes the open-world scene forgetting. 
Meanwhile, Structure-Preserving Dual Distillation~(SPDD) explicitly decomposes localization knowledge into two components for separate distillation, which includes the coarse scene layout encoded in the cluster centers activation distribution and the fine pose precision encoded in the coordinate offset residuals. This module could provide structured knowledge retention beyond standard output distillation, and prevent the student model from conflating coarse and fine signals, thereby preserving predicted camera pose precision.
When a robot arrives at a new open-world scene, the proposed continual hand-eye calibration framework can use replay samples from buffer constructed by SARS to offer information from prior scenes, and SPDD preserves the structure information by dual distillation. Subsequently, the SARS samples the newest scenes by the geometric strategy for later replay phase. 
Experiments on two public visual localization benchmarks and a simulated robotic manipulation dataset demonstrate the effectiveness of our framework. Our method achieves 74.9\% and 91.8\% average localization accuracy on i7Scenes and i12Scenes after continually learning all the scenes, outperforming the strong continual learning baseline by +8.8\% and +8.7\%, respectively. The main contributions of this paper are as follows:
\begin{itemize}
\item We propose a continual hand-eye calibration framework for open-world scene robotic manipulation, which enables a single calibration model to sequentially learn new scenes while retaining localization accuracy on previously encountered environments.
\item For ``Open-world Scene Forgetting" issue, we design a Spatial-Aware Replay Strategy sampling in joint position-orientation space, which could maximize the geometric coverage of the replay buffer and significantly decrease scene-forgetting performance.
\item For ``Pose Precision Degradation" issue, we propose a Structure-Preserving Dual Distillation mechanism, which decomposes localization knowledge for separate distillation and provides structured knowledge retention. Experimental results on multiple benchmarks confirm the effectiveness of our method in maintaining accuracy across sequentially learned scenes.
\end{itemize}

The remainder of this paper is organized as follows. Section.~\ref{sec:related} reviews related work on hand-eye calibration and continual learning. Section~\ref{sec:method} presents our proposed method and the training procedure. Section~\ref{sec:experiment} reports experimental results and ablation studies. Section~\ref{sec:conclusion} concludes the paper with a discussion of limitations and future directions.

\begin{figure*}
    \centering
    \includegraphics[width=1\linewidth]{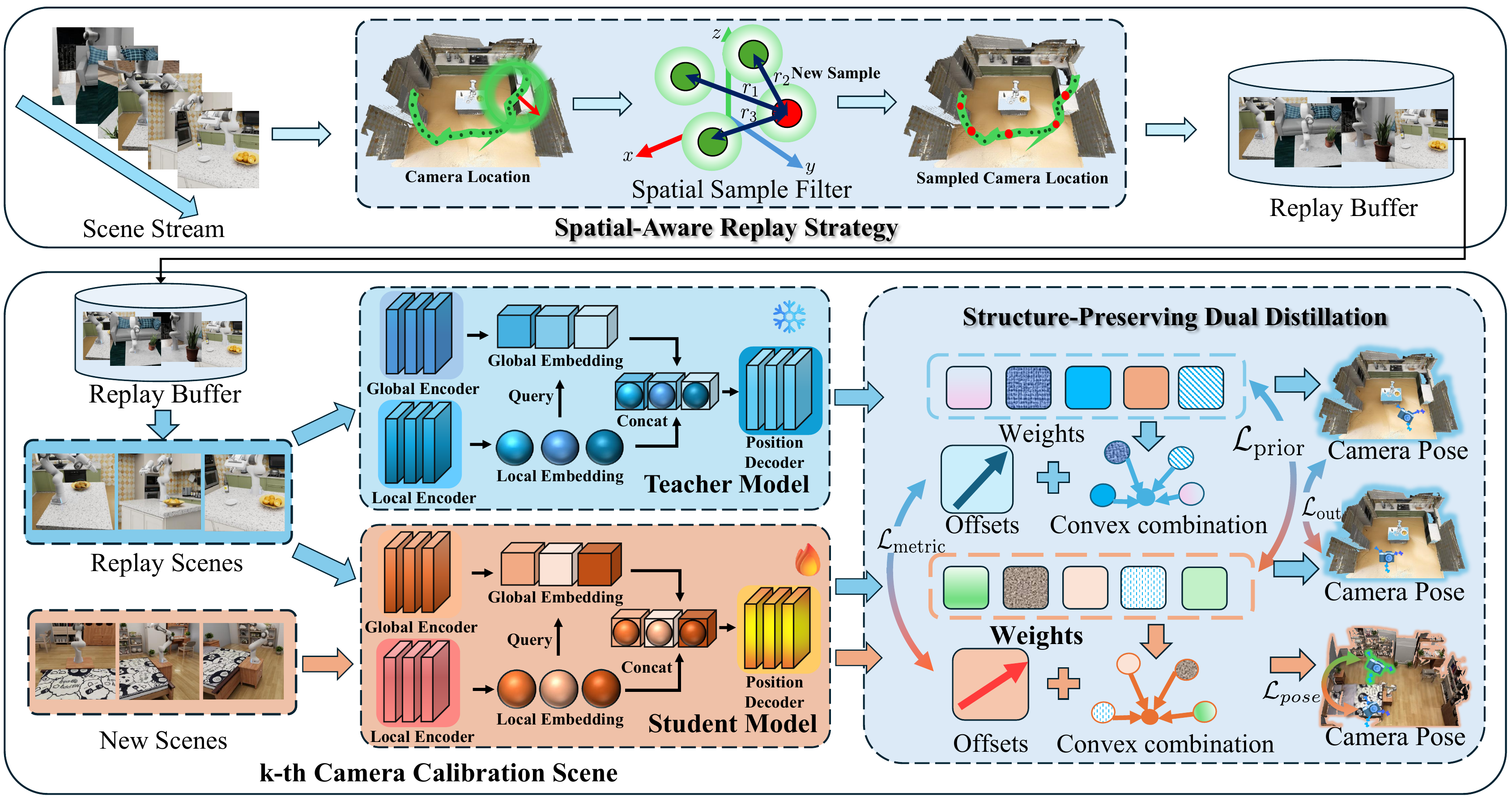}
    \caption{Overview of the proposed continual hand-eye calibration framework. \textbf{Top}: the Spatial-Aware Replay Strategy (SARS) filters incoming scene streams via Poisson disk sampling in the joint position-orientation space, retaining only geometrically diverse samples in the replay buffer. \textbf{Bottom}: during training on the $k$-th scene, a frozen teacher model and a trainable student model process replay and new scene images, respectively. The Structure-Preserving Dual Distillation (SPDD) module decomposes their outputs into cluster centers activation weights ($\mathcal{L}_{\text{prior}}$) and coordinate offsets ($\mathcal{L}_{\text{metric}}$), distilling topological and metric knowledge separately to preserve localization accuracy on previously learned scenes.}
    \label{fig:framework}
\end{figure*}

\section{Related Work}
\label{sec:related}
We briefly review prior work on hand-eye calibration and continual learning.

\subsection{Hand-Eye Calibration}
Hand-eye calibration estimates the rigid transformation between a robot and its camera, and is fundamental to robotic manipulation. Classical methods formulate this problem as $AX=XB$ and solve it by closed-form decomposition~\cite{tsai} or nonlinear optimization that minimizes $\|AX-XB\|$ or reprojection error. Recent learning-based methods directly estimate camera poses from images. CNN-based approaches detect robot keypoints and recover extrinsics via PnP, while differentiable rendering methods such as EasyHeC~\cite{easyhec} optimize camera poses by minimizing contour mismatch between rendered and real robot images. EasyHeC++~\cite{hong2024easyhec++} further improves automation with pretrained image models, and recent work also studies certifiably correct generalized robot-world and hand-eye calibration\cite{wise2026certifiably}. However, these methods are largely designed for fixed environments and typically require retraining when the scene changes.

Scene coordinate regression (SCR) offers another paradigm for visual localization by predicting per-pixel 3D scene coordinates and estimating poses with PnP-RANSAC. DSAC~\cite{dsac}, DSAC++~\cite{dsac++}, and ESAC~\cite{esac} establish and improve end-to-end SCR pipelines. ACE~\cite{ace} and GLACE~\cite{wang2024glace} remove the need for ground-truth 3D coordinates by training with reprojection loss only. HSCNet~\cite{li2020hierarchical} introduces hierarchical classification and regression, while R-SCoRe~\cite{jiang2025r}, Reloc3r\cite{dong2025reloc3r}, and scene-agnostic pose regression\cite{zheng2025scene} improve large-scene localization and cross-scene generalization. HVLF\cite{dai2025hvlf} trains the multiple scenes jointly, introducing multi-task learning~\cite{fifty2021efficiently,sun2021data} to SCR. Nevertheless, most of these methods focus on fixed-scene training or offline localization rather than continual adaptation under sequential scene changes.

\subsection{Continual Learning}
Continual learning (CL) aims to acquire knowledge from non-stationary data streams while reducing catastrophic forgetting~\cite{de2021continual,sun2024create,liang2024never}. Existing methods are commonly grouped into replay-based, regularization-based, and parameter-isolation approaches\cite{wang2024comprehensive}. Replay methods store previous samples for rehearsal, as in iCaRL~\cite{rebuffi2017icarl}. Regularization methods such as EWC~\cite{ewc} constrain updates to important parameters. Parameter-isolation methods\cite{jin2022helpful,jin2025instruction} reduce interference by assigning separate subnetworks to different tasks.

More recent replay strategies improve memory selection beyond random sampling. GSS~\cite{gss} promotes gradient diversity, MIR~\cite{mir} replays samples most affected by the current update, Co$^2$L~\cite{co2l} combines replay with contrastive learning, GEC~\cite{laigradient} introduces an explicit forgetting tolerance, and GDR~\cite{qiclass} uses leverage scores to select representative samples. However, most CL methods are developed for classification or recognition, where the preserved knowledge is mainly semantic. In contrast, continual hand-eye calibration and localization require retaining structured geometric knowledge.

Work on continual visual localization remains limited. Buff-CS~\cite{wang2021continual} first applies CL to camera relocalization and shows that SCR models suffer severe forgetting across scenes. It builds replay buffers to improve scene coverage using subscene divisions. Bahadir et al.~\cite{bahadir2024continual} extend this setting to continual hand-eye calibration by partitioning a scene into robot-centered regions and using reservoir replay. However, this setup mainly captures viewpoint variation within a single environment, rather than adaptation across distinct scenes with larger geometric changes.

Overall, existing methods still lack a unified design that jointly enforces geometrically representative memory construction and preserves pose estimation accuracy during continual adaptation, which motivates our approach.

\section{The Proposed Method}\label{sec:method}
This section presents our continual hand-eye calibration framework. We first define the problem setting in Section~\ref{sec:pd}, then introduce Spatial-Aware Replay Strategy (SARS) in Section~\ref{sec:SARS} and Structure-Preserving Dual Distillation (SPDD) in Section~\ref{sec:spdd}, and finally describe the overall training procedure in Section~\ref{sec:training}.

\subsection{Problem Definition}
\label{sec:pd}
We formulate continual hand-eye calibration as a sequential camera pose estimation task under a unified reference coordinate system.
Let $\mathcal{F}_B$, $\mathcal{F}_C$, and $\mathcal{F}_W$ denote the robot base frame, the camera optical frame, and the localization world frame, respectively. We define the world frame to coincide with the robot base frame, \textit{i.e.}, $\mathcal{F}_W \equiv \mathcal{F}_B$. Under this convention, the camera pose estimated by visual localization directly corresponds to the desired camera-to-base transformation. Given a monocular image $I \in \mathcal{I}$, the visual localization model $f_{\theta}$ predicts the camera pose $\hat{\mathbf{y}} = [\hat{\mathbf{p}}, \hat{\mathbf{q}}] \in \mathbb{R}^3 \times \mathbb{H}_1$, where $\hat{\mathbf{p}}$ and $\hat{\mathbf{q}}$ denotes the translation and unit-quaternion orientation of the camera with respect to the base frame, $f_{\theta}$ internally follows a scene coordinate regression pipeline, while the final output is the camera pose expressed in the robot base coordinate system. 
In the visual localization pipeline, the camera pose is represented as a rigid transoformation that converts a point $\mathbf{o}_i$ in the camera frame $\mathcal{F}_C$ into its corresponding 3D scene coordinate $\mathbf{x}_i$ in the world frame $\mathcal{F}_W$.
The predicted pose can be equivalently represented by the homogeneous transformation:
\begin{equation}
  \hat{\mathbf{T}}_{BC}=
\begin{bmatrix}
\mathbf{R}(\hat{\mathbf{q}}) & \hat{\mathbf{p}} \\
\mathbf{0}^{\top} & 1
\end{bmatrix}
\in \mathrm{SE}(3),
\end{equation}
which maps points from $\mathcal{F}_C$ to $\mathcal{F}_B$. 
Therefore, estimating the camera pose in the unified world/base frame is equivalent to estimating the eye-to-hand calibration result.

Open-world deployment is modeled as a sequence of scene tasks $\mathcal{T} = \{T_1, T_2, \ldots, T_N\}$ arriving sequentially. The training set of the $i$-th scene is denoted by $D_i = \{(I_j^{(i)}, \mathbf{y}_j^{(i)})\}_{j=1}^{n_i}$, where each sample consists of an RGB image and its ground-truth camera pose. During the $i$-th training stage, the model has access to the current dataset $D_i$ and a replay memory $\mathcal{M}$ that contains samples from previously learned scenes. At inference time, the model receives an image from any previously observed scene, and it predicts an accurate camera pose using a single shared parameters. This setting captures the practical requirement of open-world robotic manipulation, where environments change continuously but re-training from scratch is infeasible.
Ideally, after learning the first $i$ scenes, the model should minimize the average pose estimation error over all scenes observed so far:
\begin{equation}
\min_\theta \frac{1}{i} \sum_{k=1}^{i} \mathbb{E}_{(I,\mathbf{y}) \sim D_k} \left[ \mathcal{L}_{\text{pose}}(f_\theta(I), \mathbf{y}) \right],
\end{equation}
where $\mathcal{L}_{\text{pose}}$ denotes the GLACE\cite{wang2024glace}-style reprojection loss used by the underlying scene-coordinate-regression localizer. Under this formulation, continual hand-eye calibration must satisfy two coupled requirements. 
Firstly, the model should maintain plasticity, \textit{i.e.}, rapidly adapt to the newly arrived scene $T_i$. Secondly, it should maintain stability, i.e., preserve localization accuracy on all previously learned scenes $\{T_1, \ldots, T_{i-1}\}$. More specifically, forgetting in our problem occurs at two levels. ``Open-world Scene Forgetting" refers to losing coarse scene-level or region-level geometric association, causing the model to confuse previously learned environments. ``Pose Precision Degradation" refers to losing pose precision even when the coarse scene association is still correct, which leads to inaccurate camera-to-base transformations. The proposed SARS and SPDD are designed precisely to address these two failure modes.

\subsection{Spatial-Aware Replay Strategy}\label{sec:SARS}

In visual localization tasks, training data arrives as video streams with strong spatial correlation: consecutive frames share similar viewpoints, and adjacent segments often cover overlapping spatial regions. When conventional replay strategies such as random sampling or reservoir sampling are applied, the resulting buffer is dominated by temporally adjacent, geometrically redundant frames. This redundancy wastes the limited buffer capacity and fails to cover the spatial structure of scenes, causing the model to remember only a narrow subset of viewpoints while forgetting under-represented regions, leading to the problem of open-world scene forgetting.

To address this issue, we model the replay buffer as a sparse geometric approximation of the historical pose distribution, aiming to uniformly cover the camera pose of each scene.
Each sample $S_i$ in the buffer is defined as a tuple $S_i = \{I_i, \mathbf{p}_i, \mathbf{q}_i\}$, where $I_i$ is the image, $\mathbf{p}_i \in \mathbb{R}^3$ is the camera position, and $\mathbf{q}_i \in \mathbb{H}$ is the orientation quaternion. 
To jointly account for both translational and rotational proximity, we define a hybrid distance metric in the normalized pose space:
\begin{equation}
D(S_a, S_b) = \left\| \mathbf{p}_a - \mathbf{p}_b\right\|_2 + \lambda \cdot \text{Ang}(\mathbf{q}_a, \mathbf{q}_b),
\end{equation}
where $\text{Ang}(\mathbf{q}_a, \mathbf{q}_b) = 2 \cdot \arccos(|\mathbf{q}_a \cdot \mathbf{q}_b|)$ computes the geodesic angular difference between two quaternions, and $\lambda$ is a weighting factor that balances position and orientation contributions. This metric ensures that two samples are considered close only when they share both similar positions and similar viewing directions.

Building on this distance metric, SARS operates as an online Poisson disk sampling\cite{wei2008parallel} process with an exclusion radius $r$. For each incoming training sample $S_{new}$, the system computes its minimum distance to all existing samples in the buffer $\mathcal{M}$:
\begin{equation}
d_{min} = \min_{S_j \in \mathcal{M}} D(S_{new}, S_j).
\end{equation}
If $d_{min} < r$, the new sample is geometrically redundant with respect to the current buffer and is rejected. If $d_{min} \ge r$, the sample occupies a previously uncovered region of the pose space and is accepted. When the buffer reaches its capacity limit $N_b$, instead of random eviction, we remove the sample in the densest local neighborhood, defined as
\begin{equation}
S_{\mathrm{rm}}
=
\arg\min_{S_j \in \mathcal{M}}
\left(
\min_{S_k \in \mathcal{M}\setminus\{S_j\}} D(S_j,S_k)
\right).
\end{equation}
This density-based replacement ensures that the buffer always maintains a maximally dispersed distribution across the entire explored trajectory.

In the current implementation, SARS uses a fixed exclusion radius $r$ throughout training. Although occupancy-dependent scheduling is possible, a constant radius is sufficient in practice to suppress replay redundancy while maintaining broad geometric coverage under a bounded memory budget.

Overall, SARS produces a replay set with reduced redundancy and broader spatial-angular coverage than naive replay under the same memory budget. This is particularly beneficial for continual visual localization, where preserving viewpoint diversity is more important than preserving the temporal density of the original acquisition stream. The complete procedure is summarized in Algorithm~\ref{alg:SARS_compact}.
\begin{algorithm}[t]
\caption{SARS: Spatial-Aware Replay Buffer Update}
\label{alg:SARS_compact}
\begin{algorithmic}[1]
\REQUIRE New sample $S_{\text{new}}=(I,\mathbf{p},\mathbf{q})$, buffer $\mathcal{M}$, capacity $N_b$, weight $\lambda$, fixed radius $r$;
\ENSURE Updated buffer $\mathcal{M}$;

\STATE Use the fixed exclusion radius $r$;
\IF{$|\mathcal{M}|=0$}
    \STATE $\mathcal{M}\leftarrow \{S_{\text{new}}\}$; \;
    \RETURN $\mathcal{M}$
\ENDIF

\STATE $d_{\min}\leftarrow \min_{S_j\in\mathcal{M}}\left(\|\mathbf{p}-\mathbf{p}_j\|_2+\lambda\cdot \textsc{Ang}(\mathbf{q},\mathbf{q}_j)\right)$;
\IF{$d_{\min}<r$}
    \RETURN $\mathcal{M}$ \hspace{0.5em}\textit{/* reject redundant sample */}
\ENDIF

\IF{$|\mathcal{M}|<N_b$}
    \STATE $\mathcal{M}\leftarrow \mathcal{M}\cup\{S_{\text{new}}\}$; \;
    \RETURN $\mathcal{M}$
\ENDIF

\STATE $S_{\mathrm{rm}}\leftarrow \textsc{SelectDenseSample}(\mathcal{M})$;
\STATE $\mathcal{M}\leftarrow (\mathcal{M}\setminus\{S_{\mathrm{rm}}\})\cup\{S_{\text{new}}\}$; \;
\RETURN $\mathcal{M}$
\end{algorithmic}
\end{algorithm}

\subsection{Structure-Preserving Dual Distillation}
\label{sec:spdd}

While SARS ensures high-quality replay samples, the replay alone cannot fully prevent forgetting without appropriate training constraints. Existing continual learning methods typically apply simple output-level distillation, which treats the prediction of model as a monolithic signal. However, forgetting in visual localization manifests in two structurally distinct patterns: incorrect coarse region assignment and degraded local metric refinement. To address this, we propose a structure-aware distillation strategy that constrains coarse spatial priors and fine metric details separately.

Our distillation design builds on a GLACE-inspired cluster-center parameterization~\cite{wang2024glace}. We use a fixed set of cluster centers $\mathcal{C} = \{c_i\}_{i=1}^K \subset \mathbb{R}^3$, which is constructed once as a preprocessing step and kept unchanged throughout all continual training stages. 

For an input image $I$, the output of the localization model defines an implicit two-stage localization process. The first stage produces a logit vector $z \in \mathbb{R}^K$ whose softmax distribution encodes a probabilistic assignment over cluster centers, representing the coarse location in the scene. The second stage produces an offset tensor $\dot{d} \in \mathbb{R}^{K \times 3}$ with associated scale weights $\hat{w} \in \mathbb{R}^K$, aimint to refine the coarse cluster-center location to a precise 3D coordinate. The final scene-coordinate prediction is:
\begin{equation}
  \hat{\mathbf{x}}=\frac{\dot{d}}{\hat{w}}+\sum_{i=1}^k\frac{e^{z_i}}{\sum_je^{z_j}}c_i,
\end{equation}
\begin{equation}
w=\min(\frac{1}{Q_{\min}},\beta^{-1}\log(1+\exp(\beta\hat{w}))+\frac{1}{Q_{\max}}),
\end{equation}
where ${Q_{\min}, Q_{\max}}$ represent the lower and upper bounds of the scale, and $\beta=\frac{\log2}{1-Q_{\mathrm{max}}^{-1}}$is the parameter of softplus.

A key observation is that for any given input image, only a small number of cluster centers are geometrically relevant. The offset predictions at distant, inactive centers carry no meaningful geometric information with random noise. Blindly distilling all $K$ offset vectors would force the student to learn the noise of teacher on irrelevant centers, wasting model capacity and degrading learning efficiency on new scenes. Therefore, we restrict distillation to the set of active centers $\mathcal{A}$, defined as the top-$M_a$ centers ranked by the activation of teacher probability. All subsequent distillation losses are computed only over $\mathcal{A}$.

\noindent \textbf{Topological Prior Distillation.}
To preserve the coarse scene layout learned from previous scenes, we constrain the student distribution to match that of the teacher distribution over the active set. Specifically, we re-normalize both distributions to the active distributions over $\mathcal{A}$ and apply KL divergence:
\begin{equation}
p_{\mathcal{T},i}^{\mathcal{A}} = \frac{\exp({z_{\mathcal{T},i}^{\mathcal{A}}/ \tau})}{\Sigma_{j \in \mathcal{A}}\exp({z_{\mathcal{T},j}^{\mathcal{A}}/ \tau})}, \quad p_{\mathcal{S},i}^{\mathcal{A}} = \frac{\exp({z_{\mathcal{S},i}^{\mathcal{A}}/ \tau})}{\Sigma_{j \in \mathcal{A}}\exp({z_{\mathcal{S},j}^{\mathcal{A}}/ \tau})},
\end{equation}
\begin{equation}
\mathcal{L}_{\text{prior}} = \tau^2 \text{KL}\left(p_{\mathcal{T}}^{\mathcal{A}} \,\|\, p_{\mathcal{S}}^{\mathcal{A}}\right),
\end{equation}
where $p_{\mathcal{T}}^{\mathcal{A}}$ and $p_{\mathcal{S}}^{\mathcal{A}}$ are the re-normalized teacher and student activation probabilities over $\mathcal{A}$. This loss locks the multi-peak probabilistic prior of student for old scenes, helping to maintain the correct topological assignment, understanding which cluster center region the input belongs to.

\noindent \textbf{Metric Offset Distillation.}
To preserve fine pose precision, we constrain the coordinate offsets of student to match that of the teacher at each active cluster center. Following GLACE~\cite{wang2024glace}, the effective offset is computed in homogeneous form as $\Delta = \dot{d}/\hat{w}$. We apply an L1 loss on these residuals:
\begin{equation}
\mathcal{L}_{\text{metric}} = \left\| \frac{\dot{d}_{\mathcal{S}}[\mathcal{A}]}{\hat{w}_{\mathcal{S}}[\mathcal{A}]} - \frac{\dot{d}_{\mathcal{T}}[\mathcal{A}]}{\hat{w}_{\mathcal{T}}[\mathcal{A}]} \right\|_1.
\end{equation}
This term directly regularizes the local metric refinement stage of localization.

\noindent \textbf{Output Consistency Constraint.}
As a global safeguard, we additionally constrain the final scene coordinate prediction $\hat{y}$ in Euclidean space:
\begin{equation}
\mathcal{L}_{\text{out}} = \| \hat{\mathbf{x}}_{\mathcal{S}} - \hat{\mathbf{x}}_{\mathcal{T}} \|_1.
\end{equation}
While $\mathcal{L}_{\text{prior}}$ and $\mathcal{L}_{\text{metric}}$ target the decomposed intermediate representations, $\mathcal{L}_{\text{out}}$ ensures end-to-end consistency of the final output, preventing error accumulation from the separate distillation channels.

\noindent \textbf{Total Distillation Loss.}
The three losses are combined with weighting coefficients:
\begin{equation}
\mathcal{L}_{\text{SPDD}} = \alpha \mathcal{L}_{\text{prior}} + \beta \mathcal{L}_{\text{metric}} + \gamma \mathcal{L}_{\text{out}},
\end{equation}
where $\alpha$, $\beta$, and $\gamma$ control the relative importance of topological preservation, metric precision, and output consistency. By explicitly separating these supervisory signals, SPDD provides a structurally richer regularization signal than monolithic output distillation. When the student learns a new scene, the cluster centers distribution of old scenes is supervised by $\mathcal{L}_{\text{prior}}$, while the offset precision is maintained through $\mathcal{L}_{\text{metric}}$. The complete distillation procedure is summarized in Algorithm~\ref{alg:spdd}.
\begin{algorithm}[t]
\caption{SPDD: Distillation Loss on a Replay Batch}
\label{alg:spdd}
\begin{algorithmic}[1]
\REQUIRE Replay batch $B$, teacher $\mathcal{T}$, student $\mathcal{S}$, temperature $\tau$, active size $M_a$, weights $\alpha,\beta,\gamma$;
\ENSURE Distillation loss $\mathcal{L}_{\text{SPDD}}$;

\STATE $\mathcal{L}_\text{prior}\leftarrow 0$, \; $\mathcal{L}_\text{metric}\leftarrow 0$, \; $\mathcal{L}_\text{out}\leftarrow 0$;
\FORALL{image $I\in B$}
    \STATE $(z_{\mathcal{T}},\dot d_{\mathcal{T}},\hat w_{\mathcal{T}},\hat{\mathbf{x}}_{\mathcal{T}})\leftarrow \mathcal{T}(I)$; \hspace{0.5em}\textit{/* stop-grad */}
    \STATE $(z_{\mathcal{S}},\dot d_{\mathcal{S}},\hat w_{\mathcal{S}},\hat{\mathbf{x}}_{\mathcal{S}})\leftarrow \mathcal{S}(I)$;

    \STATE $p_{\mathcal{T}}\leftarrow \text{Softmax}(z_{\mathcal{T}}/\tau)$;
    \STATE $p_{\mathcal{S}}\leftarrow \text{Softmax}(z_{\mathcal{S}}/\tau)$;
    \STATE $\mathcal{A}\leftarrow \text{TopM}(p_{\mathcal{T}}, M_a)$;

    \STATE $p_{\mathcal{T}}^{\mathcal{A}} \leftarrow \text{Normalize}(p_{\mathcal{T}}[\mathcal{A}])$;
    \STATE $p_{\mathcal{S}}^{\mathcal{A}} \leftarrow \text{Normalize}(p_{\mathcal{S}}[\mathcal{A}])$;
    \STATE $\mathcal{L}_\text{prior} \leftarrow \mathcal{L}_\text{prior} + \tau^2\textsc{KL}\left(p_{\mathcal{T}}^{\mathcal{A}} \,\|\, p_{\mathcal{S}}^{\mathcal{A}}\right)$;

    \STATE $\Delta_{\mathcal{T}} \leftarrow \dot d_{\mathcal{T}}[\mathcal{A}] \,/\, \hat w_{\mathcal{T}}[\mathcal{A}]$;
    \STATE $\Delta_{\mathcal{S}} \leftarrow \dot d_{\mathcal{S}}[\mathcal{A}] \,/\, \hat w_{\mathcal{S}}[\mathcal{A}]$;
    \STATE $\mathcal{L}_\text{metric} \leftarrow \mathcal{L}_\text{metric} + \|\Delta_{\mathcal{S}}-\Delta_{\mathcal{T}}\|_1$;

    \STATE $\mathcal{L}_\text{out} \leftarrow \mathcal{L}_\text{out} + \|\hat{\mathbf{x}}_{\mathcal{S}}-\hat{\mathbf{x}}_{\mathcal{T}}\|_1$;
\ENDFOR

\STATE $\mathcal{L}_\text{prior} \leftarrow \mathcal{L}_\text{prior}/|B|$;
\STATE $\mathcal{L}_\text{metric} \leftarrow \mathcal{L}_\text{metric}/|B|$;
\STATE $\mathcal{L}_\text{out} \leftarrow \mathcal{L}_\text{out}/|B|$;

\RETURN $\alpha\mathcal{L}_\text{prior}+\beta\mathcal{L}_\text{metric}+\gamma\mathcal{L}_\text{out}$.
\end{algorithmic}
\end{algorithm}

\subsection{Training Procedure}
\label{sec:training}
The overall continual learning procedure proceeds sequentially over the scene sequence $\{T_1, T_2, \ldots, T_N\}$. For the first scene $T_1$, the model $f_\theta$ is trained from scratch using only $D_1$ with the standard reprojection-based pose loss $\mathcal{L}_{\text{pose}}$, following the GLACE training protocol~\cite{wang2024glace}. Simultaneously, SARS processes all incoming training samples from $D_1$ to construct the initial replay buffer $\mathcal{M}$.

For each subsequent scene $T_i$ ($i \geq 2$), the training consists of two interleaved objectives. First, a frozen copy of the model parameters from the end of scene $T_{i-1}$ is stored as the teacher model $\mathcal{T}$. The student model $\mathcal{S}$, initialized from $\mathcal{T}$, is then trained on the new scene data $D_i$ using $\mathcal{L}_{\text{pose}}$. In each training iteration, a mini-batch of replay samples is drawn from $\mathcal{M}$ and the SPDD distillation loss $\mathcal{L}_{\text{SPDD}}$ is computed against the teacher. The total training loss for scene $T_i$ is:
\begin{equation}
\mathcal{L}_{\text{total}} = \mathcal{L}_{\text{pose}} + \mathcal{L}_{\text{SPDD}}.
\end{equation}
After training on $T_i$ converges, the replay buffer $\mathcal{M}$ is updated with new samples from $D_i$ via SARS, and the trained student is frozen to serve as the teacher for the next stage. In this way, SARS and SPDD work in a complementary manner: SARS determines which past observations are retained under the memory budget, while SPDD determines how the retained knowledge is preserved during adaptation.

\section{Experiments}
\label{sec:experiment}
In this section, we evaluate the proposed continual hand-eye calibration framework on two challenging indoor localization benchmarks and own-build simulation dataset for robotic manipulation. We compare the method with several representative continual learning baselines adapted to the same visual localization backbone\cite{wang2024glace}, and perform extensive ablation studies to validate the contributions of SARS and SPDD modules.
\begin{table*}[]
\renewcommand{\tabcolsep}{1.2pt}
\caption{Median error of camera pose, and localization accuracy at the $5^\circ5\,cm$ threshold on the $\bm{i7Scenes}$ dataset. \colorbox{TableDarkGreen}{Dark green} denotes the best result under joint training, while \colorbox{TableLightGreen}{light green} denotes the best result under continual learning.}
\label{tab:i7s}
\begin{tabular}{@{}l|ccc|ccc|ccc|ccc|ccc|ccc|ccc@{}}
\toprule
\multirow{3}{*}{Scenes} & \multicolumn{21}{c}{Methods}                                                                                         \\
 &
  \multicolumn{3}{c}{Joint} &
  \multicolumn{3}{c}{Fine-tune} &
  \multicolumn{3}{c}{iCaRL\cite{rebuffi2017icarl}} &
  \multicolumn{3}{c}{Buff-CS\cite{wang2021continual}} &
  \multicolumn{3}{c}{GEC\cite{laigradient}} &
  \multicolumn{3}{c}{GDR\cite{qiclass}} &
  \multicolumn{3}{c}{Ours} \\
 &
  $\bm{t}$, cm &
  $\bm{r}$, $^\circ$ &
  Acc &
  $\bm{t}$, cm &
  $\bm{r}$, $^\circ$ &
  Acc &
  $\bm{t}$, cm &
  $\bm{r}$, $^\circ$ &
  Acc &
  $\bm{t}$, cm &
  $\bm{r}$, $^\circ$ &
  Acc &
  $\bm{t}$, cm &
  $\bm{r}$, $^\circ$ &
  Acc &
  $\bm{t}$, cm &
  $\bm{r}$, $^\circ$ &
  Acc &
  $\bm{t}$, cm &
  $\bm{r}$, $^\circ$ &
  Acc \\ \midrule
Chess &
  {\begin{tabular}{ccc} & ~{\cellcolor{TableDarkGreen}1.8}~ & \end{tabular}} &
  {\begin{tabular}{ccc} & ~{\cellcolor{TableDarkGreen}0.6}~ & \end{tabular}}  &
  {\begin{tabular}{ccc} & ~{\cellcolor{TableDarkGreen}98.8}~ & \end{tabular}}  &
  1180.7 &
  111.8 &
  0.0 &
  2.2 &
  0.8 &
  91.5 &
  {\begin{tabular}{ccc} & ~{\cellcolor{TableLightGreen}2.1}~ & \end{tabular}} &
  {\begin{tabular}{ccc} & ~{\cellcolor{TableLightGreen}0.7}~ & \end{tabular}}  &
  {\begin{tabular}{ccc} & ~{\cellcolor{TableLightGreen}92.8}~ & \end{tabular}}&
  4.2 &
  1.3 &
  59.5 &
  2.3 &
  0.8 &
  92.3 &
  {\begin{tabular}{ccc} & ~{\cellcolor{TableLightGreen}2.1}~ & \end{tabular}} &
  {\begin{tabular}{ccc} & ~{\cellcolor{TableLightGreen}0.7}~ & \end{tabular}}  &
  91.8 \\
Fire &
  {\begin{tabular}{ccc} & ~{\cellcolor{TableDarkGreen}1.7}~ & \end{tabular}} &
  {\begin{tabular}{ccc} & ~{\cellcolor{TableDarkGreen}0.7}~ & \end{tabular}}  &
  {\begin{tabular}{ccc} & ~{\cellcolor{TableDarkGreen}92.3}~ & \end{tabular}}&
  1254.8 &
  96.6 &
  0.0 &
  2.8 &
  1.1 &
  84.8 &
  2.5 &
  1.0 &
  {\begin{tabular}{ccc} & ~{\cellcolor{TableLightGreen}91.3}~ & \end{tabular}} &
  9.2 &
  3.0 &
  30.3 &
  2.6 &
  1.0 &
  89.1 &
{\begin{tabular}{ccc} & ~{\cellcolor{TableLightGreen}2.4}~ & \end{tabular}} &
  {\begin{tabular}{ccc} & ~{\cellcolor{TableLightGreen}0.9}~ & \end{tabular}}  &
  85.0 \\
Heads &
{\begin{tabular}{ccc} & ~{\cellcolor{TableDarkGreen}1.0}~ & \end{tabular}} &
  {\begin{tabular}{ccc} & ~{\cellcolor{TableDarkGreen}0.7}~ & \end{tabular}}  &
  {\begin{tabular}{ccc} & ~{\cellcolor{TableDarkGreen}99.6}~ & \end{tabular}}&
  1549.2 &
  118.8 &
  0.0 &
  2.3 &
  1.6 &
  70.5 &
  1.3 &
  0.9 &
  95.8 &
  22.1 &
  13.0 &
  3.6 &
  4.6 &
  3.1 &
  51.9 &
  {\begin{tabular}{ccc} & ~{\cellcolor{TableLightGreen}1.1}~ & \end{tabular}} &
  {\begin{tabular}{ccc} & ~{\cellcolor{TableLightGreen}0.7}~ & \end{tabular}}  &
  {\begin{tabular}{ccc} & ~{\cellcolor{TableLightGreen}97.4}~ & \end{tabular}} \\
Office &
  {\begin{tabular}{ccc} & ~{\cellcolor{TableDarkGreen}2.6}~ & \end{tabular}} &
  {\begin{tabular}{ccc} & ~{\cellcolor{TableDarkGreen}0.7}~ & \end{tabular}}  &
  {\begin{tabular}{ccc} & ~{\cellcolor{TableDarkGreen}90.1}~ & \end{tabular}}&
  671.9 &
  98.8 &
  0.0 &
  3.2 &
  1.0 &
  75.5 &
  4.0 &
  1.0 &
  64.7 &
  11.1 &
  2.7 &
  16.7 &
  4.0 &
  1.1 &
  63.9 &
    {\begin{tabular}{ccc} & ~{\cellcolor{TableLightGreen}3.1}~ & \end{tabular}} &
  {\begin{tabular}{ccc} & ~{\cellcolor{TableLightGreen}0.9}~ & \end{tabular}}  &
  {\begin{tabular}{ccc} & ~{\cellcolor{TableLightGreen}81.8}~ & \end{tabular}} \\
Pumpkin &
{\begin{tabular}{ccc} & ~{\cellcolor{TableDarkGreen}3.8}~ & \end{tabular}} &
  {\begin{tabular}{ccc} & ~{\cellcolor{TableDarkGreen}1.1}~ & \end{tabular}}  &
  {\begin{tabular}{ccc} & ~{\cellcolor{TableDarkGreen}61.1}~ & \end{tabular}}&
  1020.1 &
  126.9 &
  0.0 &
      {\begin{tabular}{ccc} & ~{\cellcolor{TableLightGreen}4.5}~ & \end{tabular}} &
  1.2 &
  54.9 &
  5.0 &
  1.3 &
  50.0 &
  35.3 &
  7.9 &
  0.5 &
  5.4 &
  1.5 &
  44.0 &
  4.6 &
  {\begin{tabular}{ccc} & ~{\cellcolor{TableLightGreen}1.1}~ & \end{tabular}}  &
  {\begin{tabular}{ccc} & ~{\cellcolor{TableLightGreen}55.4}~ & \end{tabular}}  \\
Redkitchen &
{\begin{tabular}{ccc} & ~{\cellcolor{TableDarkGreen}3.5}~ & \end{tabular}} &
  {\begin{tabular}{ccc} & ~{\cellcolor{TableDarkGreen}1.1}~ & \end{tabular}}  &
  {\begin{tabular}{ccc} & ~{\cellcolor{TableDarkGreen}70.9}~ & \end{tabular}}&
  1313.0 &
  110.8 &
  0.0 &
  5.0 &
  1.5 &
  49.7 &
  6.0 &
  1.6 &
  36.5 &
  21.6 &
  4.3 &
  4.2 &
  5.7 &
  1.7 &
  43.2 &
  {\begin{tabular}{ccc} & ~{\cellcolor{TableLightGreen}4.5}~ & \end{tabular}} &
  {\begin{tabular}{ccc} & ~{\cellcolor{TableLightGreen}1.3}~ & \end{tabular}} &
  {\begin{tabular}{ccc} & ~{\cellcolor{TableLightGreen}56.7}~ & \end{tabular}} \\
Stairs &
{\begin{tabular}{ccc} & ~{\cellcolor{TableDarkGreen}4.4}~ & \end{tabular}} &
  {\begin{tabular}{ccc} & ~{\cellcolor{TableDarkGreen}1.3}~ & \end{tabular}}  &
  {\begin{tabular}{ccc} & ~{\cellcolor{TableDarkGreen}59.7}~ & \end{tabular}}&
  16.2 &
  2.0 &
  8.3 &
  6.8 &
  1.5 &
  29.6 &
  6.9 &
  1.5 &
  31.3 &
  5.4 &
  1.3 &
  44.6 &
  6.5 &
  1.5 &
  32.7 &
  {\begin{tabular}{ccc} & ~{\cellcolor{TableLightGreen}4.6}~ & \end{tabular}} &
  {\begin{tabular}{ccc} & ~{\cellcolor{TableLightGreen}1.2}~ & \end{tabular}} &
  {\begin{tabular}{ccc} & ~{\cellcolor{TableLightGreen}56.2}~ & \end{tabular}} \\ \midrule
Accuracy $(\uparrow)$ &
  \multicolumn{3}{c|}{{\begin{tabular}{ccc} & ~{\cellcolor{TableDarkGreen}81.7}~ & \end{tabular}}} &
  \multicolumn{3}{c|}{1.12} &
  \multicolumn{3}{c|}{65.2} &
  \multicolumn{3}{c|}{66.1} &
  \multicolumn{3}{c|}{26.5} &
  \multicolumn{3}{c|}{59.6} &
  \multicolumn{3}{c}{ {\begin{tabular}{ccc} & ~{\cellcolor{TableLightGreen}74.9}~ & \end{tabular}}} \\ 
  TFR $(\downarrow)$&
  \multicolumn{3}{c|}{-} &
  \multicolumn{3}{c|}{-} &
  \multicolumn{3}{c|}{11.37} &
  \multicolumn{3}{c|}{13.03} &
  \multicolumn{3}{c|}{62.67} &
  \multicolumn{3}{c|}{12.25} &
  \multicolumn{3}{c}{ {\begin{tabular}{ccc} & ~{\cellcolor{TableLightGreen}6.65}~ & \end{tabular}}} \\
  \bottomrule
\end{tabular}
\end{table*}
\begin{table*}[]
\renewcommand{\tabcolsep}{0.8pt}
\caption{Median error of camera pose, and localization accuracy at the $5^\circ5\,cm$ threshold on the $\bm{i12Scenes}$ dataset. \colorbox{TableDarkGreen}{Dark green} denotes the best result under joint training, while \colorbox{TableLightGreen}{light green} denotes the best result under continual learning.}
\label{tab:i12s}
\begin{tabular}{@{}l|ccc|ccc|ccc|ccc|ccc|ccc|ccc@{}}
\toprule
\multirow{3}{*}{Scenes} & \multicolumn{21}{c}{Methods}                                                                                         \\
 &
  \multicolumn{3}{c}{Joint} &
  \multicolumn{3}{c}{Fine-tune} &
  \multicolumn{3}{c}{iCaRL\cite{rebuffi2017icarl}} &
  \multicolumn{3}{c}{Buff-CS\cite{wang2021continual}} &
  \multicolumn{3}{c}{GEC\cite{laigradient}} &
  \multicolumn{3}{c}{GDR\cite{qiclass}} &
  \multicolumn{3}{c}{Ours} \\
 &
  $\bm{t}$, cm &
  $\bm{r}$, $^\circ$ &
  Acc &
  $\bm{t}$, cm &
  $\bm{r}$, $^\circ$ &
  Acc &
  $\bm{t}$, cm &
  $\bm{r}$, $^\circ$ &
  Acc &
  $\bm{t}$, cm &
  $\bm{r}$, $^\circ$ &
  Acc &
  $\bm{t}$, cm &
  $\bm{r}$, $^\circ$ &
  Acc &
  $\bm{t}$, cm &
  $\bm{r}$, $^\circ$ &
  Acc &
  $\bm{t}$, cm &
  $\bm{r}$, $^\circ$ &
  Acc \\ \midrule
Kitchen-1 &
  {\begin{tabular}{ccc} & ~{\cellcolor{TableDarkGreen}0.8}~ & \end{tabular}} &
  {\begin{tabular}{ccc} & ~{\cellcolor{TableDarkGreen}0.5}~ & \end{tabular}} &
  {\begin{tabular}{ccc} & ~{\cellcolor{TableDarkGreen}99.2}~ & \end{tabular}} &
  1937.2 &
  129.5 &
  0.0 &
  1.6 &
  0.9 &
  93.8 &
  1.5 &
  0.9 &
  88.8 &
  243.4 &
  20.2 &
  0.0 &
  {\begin{tabular}{ccc} & ~{\cellcolor{TableLightGreen}1.3}~ & \end{tabular}} &
  {\begin{tabular}{ccc} & ~{\cellcolor{TableLightGreen}0.8}~ & \end{tabular}} &
  {\begin{tabular}{ccc} & ~{\cellcolor{TableLightGreen}93.8}~ & \end{tabular}} &
  1.5 &
  {\begin{tabular}{ccc} & ~{\cellcolor{TableLightGreen}0.8}~ & \end{tabular}} & 
  92.7 \\
Living-1 &
  {\begin{tabular}{ccc} & ~{\cellcolor{TableDarkGreen}1.1}~ & \end{tabular}} &
  {\begin{tabular}{ccc} & ~{\cellcolor{TableDarkGreen}0.4}~ & \end{tabular}} &
  {\begin{tabular}{ccc} & ~{\cellcolor{TableDarkGreen}100.0}~ & \end{tabular}} &
  1495.4 &
  114.3 &
  0.0 &
  1.8 &
  0.7 &
  90.9 &
  {\begin{tabular}{ccc} & ~{\cellcolor{TableLightGreen}1.4}~ & \end{tabular}} &
  0.6 &
  95.7 &
  16.1 &
  5.5 &
  17.8 &
  {\begin{tabular}{ccc} & ~{\cellcolor{TableLightGreen}1.4}~ & \end{tabular}} &
  {\begin{tabular}{ccc} & ~{\cellcolor{TableLightGreen}0.5}~ & \end{tabular}} &
  95.9 &
  1.5 &
  0.6 &
  {\begin{tabular}{ccc} & ~{\cellcolor{TableLightGreen}97.2}~ & \end{tabular}}  \\
Bed &
  {\begin{tabular}{ccc} & ~{\cellcolor{TableDarkGreen}1.1}~ & \end{tabular}} &
  {\begin{tabular}{ccc} & ~{\cellcolor{TableDarkGreen}0.5}~ & \end{tabular}} &
  {\begin{tabular}{ccc} & ~{\cellcolor{TableDarkGreen}99.5}~ & \end{tabular}} &
  1215.3 &
  121.9 &
  0.0 &
  4.5 &
  1.9 &
  53.9 &
  1.9 &
  0.8 &
  {\begin{tabular}{ccc} & ~{\cellcolor{TableLightGreen}95.6}~ & \end{tabular}} &
  135.8 &
  37.3 &
  3.9 &
  2.2 &
  1.0 &
  92.2 &
  {\begin{tabular}{ccc} & ~{\cellcolor{TableLightGreen}1.4}~ & \end{tabular}} &
  {\begin{tabular}{ccc} & ~{\cellcolor{TableLightGreen}0.6}~ & \end{tabular}} &
  93.6 \\
Kitchen-2 &
  {\begin{tabular}{ccc} & ~{\cellcolor{TableDarkGreen}1.0}~ & \end{tabular}} &
  {\begin{tabular}{ccc} & ~{\cellcolor{TableDarkGreen}0.5}~ & \end{tabular}} &
  {\begin{tabular}{ccc} & ~{\cellcolor{TableDarkGreen}99.9}~ & \end{tabular}} &
  1074.9 &
  130.9 &
  0.0 &
  2.1 &
  1.0 &
  87.6 &
  2.1 &
  0.9 &
  81.0 &
  395.1 &
  18.1 &
  0.0 &
  2.4 &
  1.0 &
  81.0 &
  {\begin{tabular}{ccc} & ~{\cellcolor{TableLightGreen}1.1}~ & \end{tabular}} &
  {\begin{tabular}{ccc} & ~{\cellcolor{TableLightGreen}0.5}~ & \end{tabular}} &
  {\begin{tabular}{ccc} & ~{\cellcolor{TableLightGreen}97.6}~ & \end{tabular}}  \\
Living-2 &
  {\begin{tabular}{ccc} & ~{\cellcolor{TableDarkGreen}1.0}~ & \end{tabular}} &
  {\begin{tabular}{ccc} & ~{\cellcolor{TableDarkGreen}0.4}~ & \end{tabular}} &
  {\begin{tabular}{ccc} & ~{\cellcolor{TableDarkGreen}98.9}~ & \end{tabular}} &
  1691.3 &
  126.4 &
  0.0 &
  6.1 &
  2.4 &
  41.8 &
  1.7 &
  0.7 &
  95.7 &
  351.6 &
  43.2 &
  0.9 &
  2.0 &
  0.7 &
  90.3 &
  {\begin{tabular}{ccc} & ~{\cellcolor{TableLightGreen}1.5}~ & \end{tabular}} &
  {\begin{tabular}{ccc} & ~{\cellcolor{TableLightGreen}0.6}~ & \end{tabular}} &
  {\begin{tabular}{ccc} & ~{\cellcolor{TableLightGreen}98.9}~ & \end{tabular}}  \\
Luke &
  {\begin{tabular}{ccc} & ~{\cellcolor{TableDarkGreen}1.3}~ & \end{tabular}} &
  {\begin{tabular}{ccc} & ~{\cellcolor{TableDarkGreen}0.6}~ & \end{tabular}} &
  {\begin{tabular}{ccc} & ~{\cellcolor{TableDarkGreen}100.0}~ & \end{tabular}} &
  1299.4 &
  117.9 &
  0.0 &
  6.2 &
  2.2 &
  42.6 &
  3.3 &
  1.4 &
  67.3 &
  58.9 &
  19.0 &
  5.4 &
  3.5 &
  1.5 &
  65.7 &
  {\begin{tabular}{ccc} & ~{\cellcolor{TableLightGreen}2.3}~ & \end{tabular}} &
  {\begin{tabular}{ccc} & ~{\cellcolor{TableLightGreen}1.0}~ & \end{tabular}} &
  {\begin{tabular}{ccc} & ~{\cellcolor{TableLightGreen}86.5}~ & \end{tabular}}  \\
Gate362 &
  {\begin{tabular}{ccc} & ~{\cellcolor{TableDarkGreen}1.0}~ & \end{tabular}} &
  {\begin{tabular}{ccc} & ~{\cellcolor{TableDarkGreen}0.4}~ & \end{tabular}} &
  {\begin{tabular}{ccc} & ~{\cellcolor{TableDarkGreen}98.0}~ & \end{tabular}} &
  888.2 &
  127.8 &
  0.0 &
  {\begin{tabular}{ccc} & ~{\cellcolor{TableLightGreen}1.5}~ & \end{tabular}} &
  {\begin{tabular}{ccc} & ~{\cellcolor{TableLightGreen}0.6}~ & \end{tabular}} &
  {\begin{tabular}{ccc} & ~{\cellcolor{TableLightGreen}100.0}~ & \end{tabular}} &
  2.1 &
  0.9 &
  87.8 &
  36.7 &
  4.6 &
  0.8 &
  2.0 &
  0.9 &
  87.3 &
  1.7 &
  0.8 &
  97.9 \\
Gate381 &
  {\begin{tabular}{ccc} & ~{\cellcolor{TableDarkGreen}1.5}~ & \end{tabular}} &
  {\begin{tabular}{ccc} & ~{\cellcolor{TableDarkGreen}0.7}~ & \end{tabular}} &
  {\begin{tabular}{ccc} & ~{\cellcolor{TableDarkGreen}76.1}~ & \end{tabular}} &
  603.2 &
  119.1 &
  0.0 &
  4.9 &
  2.2 &
  51.3 &
  3.3 &
  {\begin{tabular}{ccc} & ~{\cellcolor{TableLightGreen}1.5}~ & \end{tabular}} &
  68.6 &
  13.2 &
  5.3 &
  16.8 &
  5.7 &
  2.7 &
  44.8 &
  {\begin{tabular}{ccc} & ~{\cellcolor{TableLightGreen}3.2}~ & \end{tabular}} &
  {\begin{tabular}{ccc} & ~{\cellcolor{TableLightGreen}1.5}~ & \end{tabular}} &
  {\begin{tabular}{ccc} & ~{\cellcolor{TableLightGreen}71.9}~ & \end{tabular}} \\
Lounge &
  {\begin{tabular}{ccc} & ~{\cellcolor{TableDarkGreen}1.3}~ & \end{tabular}} &
  {\begin{tabular}{ccc} & ~{\cellcolor{TableDarkGreen}0.4}~ & \end{tabular}} &
  {\begin{tabular}{ccc} & ~{\cellcolor{TableDarkGreen}99.1}~ & \end{tabular}} &
  1525.1 &
  123.2 &
  0.0 &
  4.0 &
  1.2 &
  56.6 &
  2.5 &
  {\begin{tabular}{ccc} & ~{\cellcolor{TableLightGreen}0.8}~ & \end{tabular}} &
  87.8 &
  16.7 &
  6.1 &
  5.8 &
  3.2 &
  1.1 &
  73.1 &
  {\begin{tabular}{ccc} & ~{\cellcolor{TableLightGreen}2.2}~ & \end{tabular}} &
  {\begin{tabular}{ccc} & ~{\cellcolor{TableLightGreen}0.8}~ & \end{tabular}} &
  {\begin{tabular}{ccc} & ~{\cellcolor{TableLightGreen}93.9}~ & \end{tabular}} \\
Manolis &
  {\begin{tabular}{ccc} & ~{\cellcolor{TableDarkGreen}1.2}~ & \end{tabular}} &
  {\begin{tabular}{ccc} & ~{\cellcolor{TableDarkGreen}0.5}~ & \end{tabular}} &
  {\begin{tabular}{ccc} & ~{\cellcolor{TableDarkGreen}97.5}~ & \end{tabular}} &
  1156.5 &
  120.5 &
  0.0 &
  3.7 &
  1.5 &
  66.2 &
  3.5 &
  1.5 &
  66.8 &
  642.2 &
  53.9 &
  6.4 &
  3.5 &
  1.5 &
  63.7 &
  {\begin{tabular}{ccc} & ~{\cellcolor{TableLightGreen}2.5}~ & \end{tabular}} &
  {\begin{tabular}{ccc} & ~{\cellcolor{TableLightGreen}1.0}~ & \end{tabular}} &
  {\begin{tabular}{ccc} & ~{\cellcolor{TableLightGreen}85.8}~ & \end{tabular}} \\
Floor.5a &
  {\begin{tabular}{ccc} & ~{\cellcolor{TableDarkGreen}1.5}~ & \end{tabular}} &
  {\begin{tabular}{ccc} & ~{\cellcolor{TableDarkGreen}0.6}~ & \end{tabular}} &
  {\begin{tabular}{ccc} & ~{\cellcolor{TableDarkGreen}95.2}~ & \end{tabular}} &
  837.1 &
  104.0 &
  0.0 &
  11.2 &
  3.5 &
  20.5 &
  3.8 &
  1.5 &
  62.6 &
  381.9 &
  56.1 &
  4.8 &
  4.2 &
  1.6 &
  57.5 &
  {\begin{tabular}{ccc} & ~{\cellcolor{TableLightGreen}2.7}~ & \end{tabular}} &
  {\begin{tabular}{ccc} & ~{\cellcolor{TableLightGreen}1.1}~ & \end{tabular}} &
  {\begin{tabular}{ccc} & ~{\cellcolor{TableLightGreen}85.7}~ & \end{tabular}} \\
Floor.5b &
  {\begin{tabular}{ccc} & ~{\cellcolor{TableDarkGreen}1.5}~ & \end{tabular}} &
  {\begin{tabular}{ccc} & ~{\cellcolor{TableDarkGreen}0.5}~ & \end{tabular}} &
  {\begin{tabular}{ccc} & ~{\cellcolor{TableDarkGreen}100.0}~ & \end{tabular}} &
  2.8 &
  0.9 &
  75.1 &
  1.7 &
  0.5 &
  95.6 &
  1.5 &
  0.5 &
  99.8 &
  1.5 &
  0.4 &
  99.8 &
  1.7 &
  0.5 &
  99.3 &
  {\begin{tabular}{ccc} & ~{\cellcolor{TableLightGreen}1.6}~ & \end{tabular}} &
  {\begin{tabular}{ccc} & ~{\cellcolor{TableLightGreen}0.5}~ & \end{tabular}} &
  {\begin{tabular}{ccc} & ~{\cellcolor{TableLightGreen}99.8}~ & \end{tabular}}  \\ \midrule
Accuracy $(\uparrow)$ &
  \multicolumn{3}{c|}{ {\begin{tabular}{ccc} & ~{\cellcolor{TableDarkGreen}96.9}~ & \end{tabular}}} &
  \multicolumn{3}{c|}{6.2} &
  \multicolumn{3}{c|}{66.0} &
  \multicolumn{3}{c|}{83.1} &
  \multicolumn{3}{c|}{13.5} &
  \multicolumn{3}{c|}{78.7} &
  \multicolumn{3}{c}{ {\begin{tabular}{ccc} & ~{\cellcolor{TableLightGreen}91.8}~ & \end{tabular}}} \\  
  TFR $(\downarrow)$&
  \multicolumn{3}{c|}{-} &
  \multicolumn{3}{c|}{-} &
  \multicolumn{3}{c|}{33.3} &
  \multicolumn{3}{c|}{15.7} &
  \multicolumn{3}{c|}{88.9} &
  \multicolumn{3}{c|}{21.9} &
  \multicolumn{3}{c}{ {\begin{tabular}{ccc} & ~{\cellcolor{TableLightGreen}8.1}~ & \end{tabular}}} \\
  \bottomrule
\end{tabular}
\end{table*}

\begin{table*}[]
\renewcommand{\tabcolsep}{1.2pt}
\caption{Median error of camera pose, and localization accuracy at the $1^\circ1\,cm$ threshold on the $\bm{Simulation}$ dataset. \colorbox{TableDarkGreen}{Dark green} denotes the best result under joint training, while \colorbox{TableLightGreen}{light green} denotes the best result under continual learning.}
\label{tab:sim}
\begin{tabular}{@{}l|ccc|ccc|ccc|ccc|ccc|ccc|ccc@{}}
\toprule
\multirow{3}{*}{Scenes} & \multicolumn{21}{c}{Methods}                                                                                         \\
 &
  \multicolumn{3}{c}{Joint} &
  \multicolumn{3}{c}{Fine-tune} &
  \multicolumn{3}{c}{iCaRL\cite{rebuffi2017icarl}} &
  \multicolumn{3}{c}{Buff-CS\cite{wang2021continual}} &
  \multicolumn{3}{c}{GEC\cite{laigradient}} &
  \multicolumn{3}{c}{GDR\cite{qiclass}} &
  \multicolumn{3}{c}{Ours} \\
 &
  $\bm{t}$, cm &
  $\bm{r}$, $^\circ$ &
  Acc &
  $\bm{t}$, cm &
  $\bm{r}$, $^\circ$ &
  Acc &
  $\bm{t}$, cm &
  $\bm{r}$, $^\circ$ &
  Acc &
  $\bm{t}$, cm &
  $\bm{r}$, $^\circ$ &
  Acc &
  $\bm{t}$, cm &
  $\bm{r}$, $^\circ$ &
  Acc &
  $\bm{t}$, cm &
  $\bm{r}$, $^\circ$ &
  Acc &
  $\bm{t}$, cm &
  $\bm{r}$, $^\circ$ &
  Acc \\ \midrule
Bedroom &
  {\begin{tabular}{ccc} & ~{\cellcolor{TableDarkGreen}0.6}~ & \end{tabular}} &
  {\begin{tabular}{ccc} & ~{\cellcolor{TableDarkGreen}0.2}~ & \end{tabular}} &
  {\begin{tabular}{ccc} & ~{\cellcolor{TableDarkGreen}97.3}~ & \end{tabular}} &
  479.9 &
  111.3 &
  0.0 &
  1.1 &
  0.2 &
  40.2 &
  \multicolumn{3}{c|}{-} &
  0.5 &
  {\begin{tabular}{ccc} & ~{\cellcolor{TableLightGreen}0.1}~ & \end{tabular}}&
  91.7 &
  0.7 &
  0.2 &
  91.2 &
  0.6 &
  {\begin{tabular}{ccc} & ~{\cellcolor{TableLightGreen}0.1}~ & \end{tabular}} &
  {\begin{tabular}{ccc} & ~{\cellcolor{TableLightGreen}96.2}~ & \end{tabular}}\\
Kitchen &
  {\begin{tabular}{ccc} & ~{\cellcolor{TableDarkGreen}0.5}~ & \end{tabular}} &
  {\begin{tabular}{ccc} & ~{\cellcolor{TableDarkGreen}0.1}~ & \end{tabular}} &
  {\begin{tabular}{ccc} & ~{\cellcolor{TableDarkGreen}99.8}~ & \end{tabular}} &
  774.6 &
  109.8 &
  0.0 &
  0.7 &
  0.2 &
  68.8 &
  \multicolumn{3}{c|}{-} &
  0.6 &
  {\begin{tabular}{ccc} & ~{\cellcolor{TableLightGreen}0.1}~ & \end{tabular}} &
  94.7 &
  0.6 &
  {\begin{tabular}{ccc} & ~{\cellcolor{TableLightGreen}0.1}~ & \end{tabular}} &
  94.7 &
    {\begin{tabular}{ccc} & ~{\cellcolor{TableLightGreen}0.4}~ & \end{tabular}} &
    {\begin{tabular}{ccc} & ~{\cellcolor{TableLightGreen}0.1}~ & \end{tabular}} &
    {\begin{tabular}{ccc} & ~{\cellcolor{TableLightGreen}99.4}~ & \end{tabular}}  \\
Living &
{\begin{tabular}{ccc} & ~{\cellcolor{TableDarkGreen}0.2}~ & \end{tabular}} &
  {\begin{tabular}{ccc} & ~{\cellcolor{TableDarkGreen}0.1}~ & \end{tabular}} &
  {\begin{tabular}{ccc} & ~{\cellcolor{TableDarkGreen}100.0}~ & \end{tabular}} &

  0.3 &
  0.1 &
  100.0 &
  0.4 &
  0.1 &
  99.6 &
  \multicolumn{3}{c|}{-} &
  0.4 &
  {\begin{tabular}{ccc} & ~{\cellcolor{TableLightGreen}0.1}~ & \end{tabular}} &
  96.3 &
  0.4 &
  {\begin{tabular}{ccc} & ~{\cellcolor{TableLightGreen}0.1}~ & \end{tabular}} &
  {\begin{tabular}{ccc} & ~{\cellcolor{TableLightGreen}100.0}~ & \end{tabular}} &
  {\begin{tabular}{ccc} & ~{\cellcolor{TableLightGreen}0.3}~ & \end{tabular}} &
  {\begin{tabular}{ccc} & ~{\cellcolor{TableLightGreen}0.1}~ & \end{tabular}} &
  99.6 \\ \midrule
Accuracy $(\uparrow)$ &
  \multicolumn{3}{c|}{{\begin{tabular}{ccc} & ~{\cellcolor{TableDarkGreen}99.0}~ & \end{tabular}}} &
  \multicolumn{3}{c|}{33.3} &
  \multicolumn{3}{c|}{69.5} &
  \multicolumn{3}{c|}{-} &
  \multicolumn{3}{c|}{94.2} &
  \multicolumn{3}{c|}{95.3} &
  \multicolumn{3}{c}{ {\begin{tabular}{ccc} & ~{\cellcolor{TableLightGreen}98.4}~ & \end{tabular}}} \\ 
  TFR $(\downarrow)$&
  \multicolumn{3}{c|}{-} &
  \multicolumn{3}{c|}{-} &
  \multicolumn{3}{c|}{44.9} &
  \multicolumn{3}{c|}{-} &
  \multicolumn{3}{c|}{5.1} &
  \multicolumn{3}{c|}{6.5} &
  \multicolumn{3}{c}{ {\begin{tabular}{ccc} & ~{\cellcolor{TableLightGreen}1.6}~ & \end{tabular}}} \\
  \bottomrule
\end{tabular}
\end{table*}
\subsection{Experimental Setup}
\begin{figure*}[htp]
    \centering
    \includegraphics[width=\textwidth]{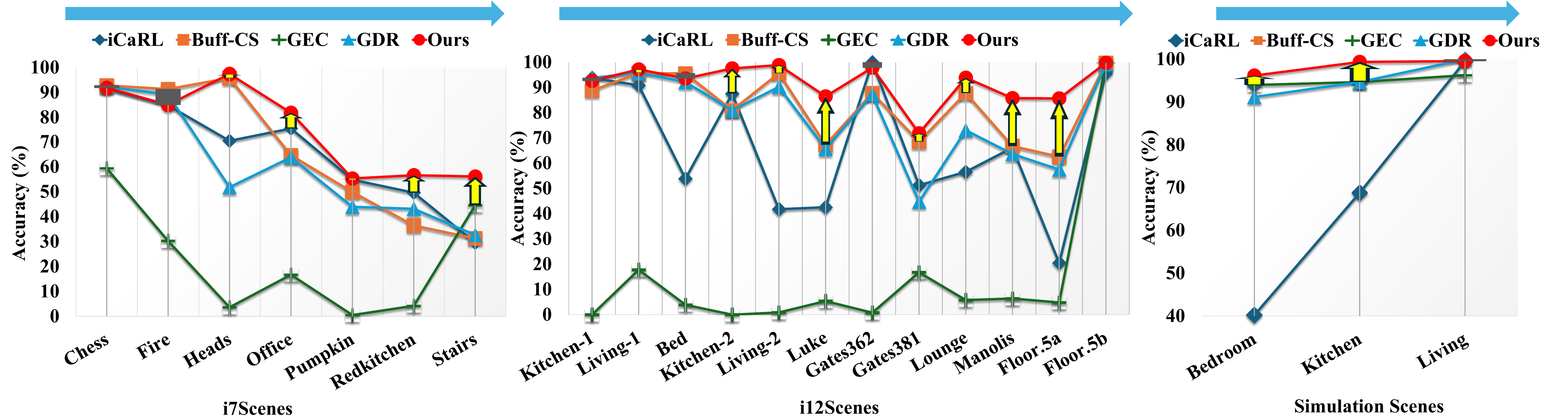}
    \caption{Per-scene localization accuracy comparison across all three datasets. Each group shows the final accuracy of all continual learning methods on individual scenes after training on the complete scene sequence. Yellow arrows highlight scenes where our method achieves the largest improvement over baselines.}
    \label{fig:perscene}
\end{figure*}
\noindent \textbf{Datasets.}
We conduct experiments on two public continual-localization benchmarks based on \textbf{7Scenes}~\cite{7scenes} and \textbf{12Scenes}~\cite{valentin2016learning}.
Following~\cite{wang2021continual}, \textbf{i7S} and \textbf{i12S} are obtained by registering the original \textbf{7Scenes} and \textbf{12Scenes} sequences into unified coordinate systems, yielding two multi-scene continual-learning environments. 
\textbf{i7S} contains seven indoor scenes and covers approximately $125\,\mathrm{m}^3$, whereas \textbf{i12S} contains 12 scenes across four larger indoor environments and covers approximately $520\,\mathrm{m}^3$.

To validate our method on robotic manipulation scenarios, we additionally construct a \textbf{Sim} dataset using Isaac Sim. We utilize three photorealistic indoor environments (Bedroom, Kitchen, and Living room) equipped with Franka. By constructing camera trajectories with controlled observation directions, we collect approximately 10,000 RGB-D frames with ground-truth camera poses across the three scenes for continual learning evaluation.

\begin{figure*}[htp]
    \centering
    \includegraphics[width=1\textwidth]{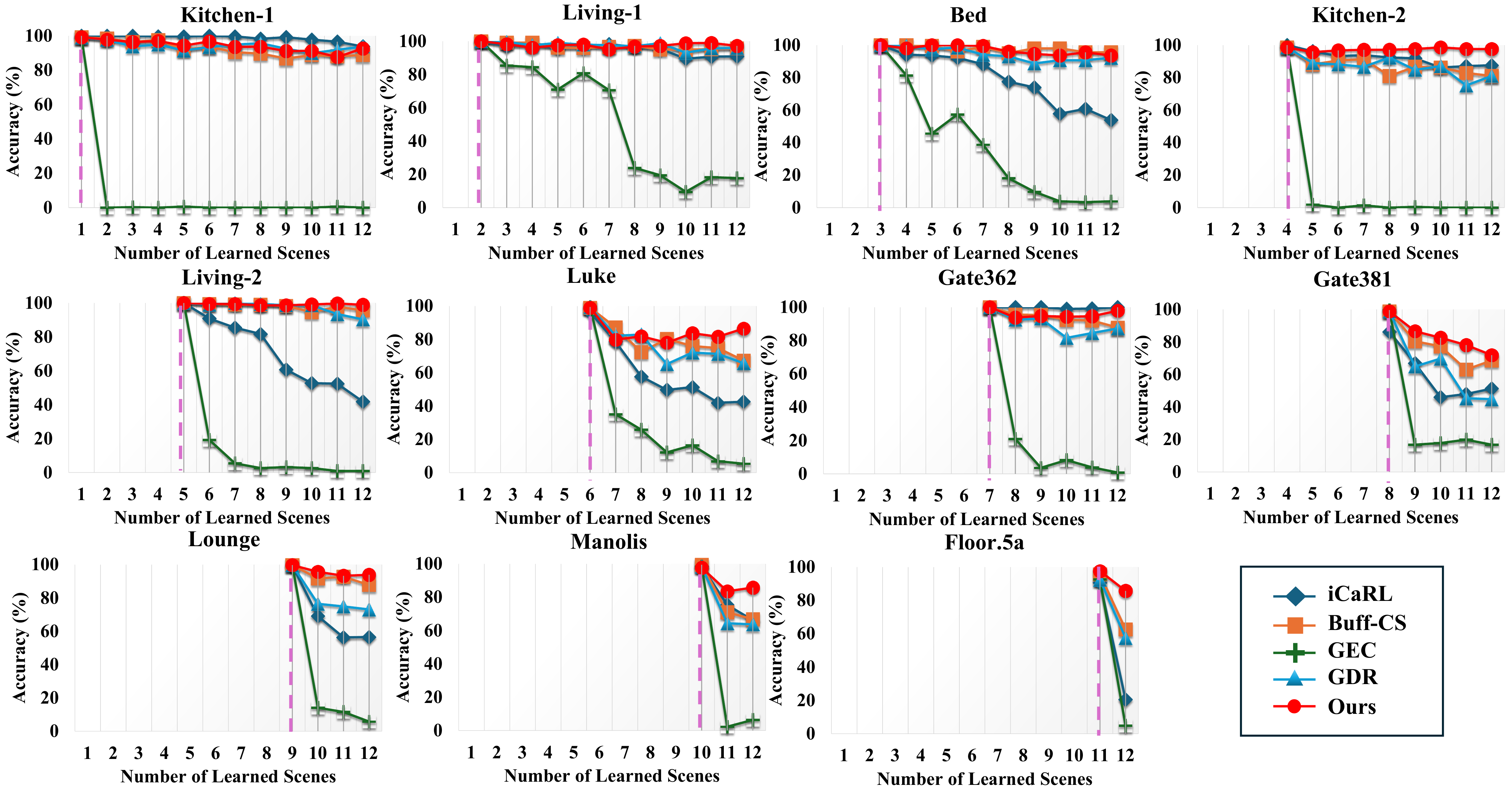}
    \caption{Task forgetting rate comparisons on the i12S dataset.}
    \label{fig:i12stfr}
\end{figure*}
\begin{figure}[htp]
    \centering
    \includegraphics[width=0.47\textwidth]{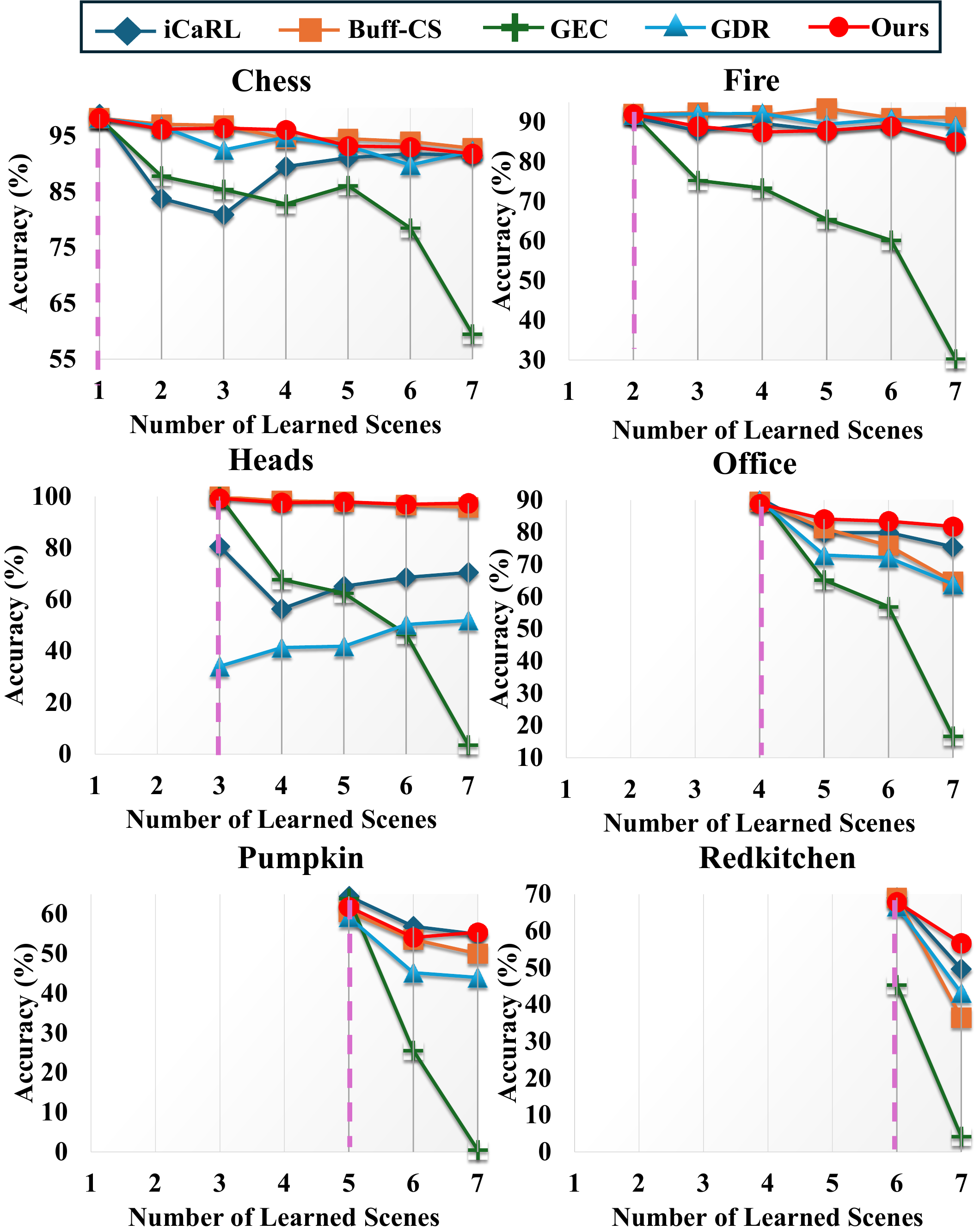}
    \caption{Task forgetting rate comparisons on the i7S dataset.}
    \label{fig:i7stfr}
\end{figure}

\noindent \textbf{Baselines.}
We compare the proposed method with six baselines. Joint trains a single model on all scenes simultaneously, serving as the performance upper bound. 
Fine-tune represents training sequentially without any anti-forgetting mechanism, providing the lower bound. 
iCaRL~\cite{rebuffi2017icarl} combines exemplar replay with knowledge distillation using herding-based sample selection. Buff-CS~\cite{wang2021continual} constructs a replay buffer that explicitly targets spatial scene coverage by selecting images spanning different spatial regions; we re-implement its core replay strategy on the GLACE backbone for a fair comparison using the preprocessed dataset provided by the original authors. However, on our own-built \textbf{Sim} dataset, the preprocessed data is not available, so the Buff-CS results are not reported for \textbf{Sim} dataset.
GEC~\cite{laigradient} introduces gradient-guided $\epsilon$-forgetting tolerance constraints to adaptively balance stability and plasticity. 
GDR~\cite{qiclass} uses SVD-based leverage scores to select globally representative replay samples.

\noindent \textbf{Evaluation Metrics.}
Following~\cite{wang2021continual}, we use localization accuracy as the primary metric, defined as the percentage of query images whose translation and rotation errors are below $5$ cm and $5^\circ$ on i7S/i12S, and below $1$ cm and $1^\circ$ on Sim. We additionally report the median translation error, the median rotation error, the final accuracy after learning the complete scene sequence, and the total forgetting rate (TFR)~\cite{diaz2018don}, which measures the average drop from the best historical accuracy of each scene to its final accuracy. Higher accuracy and lower TFR indicate better continual adaptation.

\noindent \textbf{Implementation Details.}
All methods are built on the GLACE~\cite{wang2024glace} as the visual localization backbone and follow its training protocol. We train with 6 GPUs and set the iteration steps to 15K per scene on i7S and i12S datasets, and 5K iterations on Sim. We use the AdamW optimizer with an initial learning rate of 0.005. For our method, the replay buffer capacity $N_b$ is set to 10\% samples per scene, the exclusion radius $r$ is 0.5 in the normalized pose space, and the distance weighting factor $\lambda$ is 1.0. All baseline methods follow the same training setup. For SPDD, the distillation loss weights are set to $\alpha = 1.0$, $\beta = 1.0$, $\gamma = 1.0$, the softmax temperature $\tau = 2.0$, and the active cluster centers size $M_a = 50$.

\subsection{Comparison with State-of-the-Arts}

Tables~\ref{tab:i7s},~\ref{tab:i12s}, and~\ref{tab:sim} report the per-scene median pose error and localization accuracy for all methods on the i7S, i12S, and Sim datasets, respectively.

\noindent \textbf{Upper and lower bounds.}
Joint training establishes a strong upper bound, achieving 81.7\% on i7S, 96.9\% on i12S, and 99.0\% on Sim. 
In contrast, naive fine-tuning degrades to 1.12\%, 6.2\%, and 33.3\%, respectively, showing that SCR-based calibration suffers severe forgetting under sequential scene adaptation.

\noindent \textbf{Baseline analysis.}
Among continual learning baselines, methods directly borrowed from classification show clear limitations in the localization setting. 
Buff-CS is competitive on i7S and i12S because it explicitly promotes scene coverage during buffer construction.
GDR is relatively strong on Sim, suggesting that globally representative replay already helps in the smaller three-scene setting.
By contrast, GEC performs poorly on i7S and i12S, with final accuracies of only 26.5\% and 13.5\%, respectively, indicating that a generic gradient-constrained strategy is insufficient for preserving the structured geometric knowledge required in pose estimation.
iCaRL remains a reasonable replay-based baseline but shows large cross-scene variance, especially on the larger i12S benchmark.

\noindent \textbf{Results on i7S.}
On the i7S dataset (Table~\ref{tab:i7s}), our method achieves a final accuracy of 74.9\%, substantially outperforming iCaRL (65.2\%), Buff-CS (66.1\%), GDR (59.6\%), and GEC (26.5\%). The corresponding TFR is 6.65, which is lower than all competing continual-learning baselines, including 11.37 for iCaRL, 13.03 for Buff-CS, and 12.25 for GDR. Relative to Buff-CS, which is the strongest baseline in final accuracy on this dataset, our method improves the final accuracy by 8.8 percentage points while reducing forgetting by roughly half. The remaining gap to Joint is only 6.8 points. These results suggest that improving both replay coverage and replay-time knowledge preservation is beneficial even in the shorter seven-scene sequence.

As shown in Fig.~\ref{fig:perscene}~(left), our method achieves the highest or near-highest accuracy on every scene. The most striking gains occur on the difficult scenes, such as \emph{Office}, \emph{Redkitchen}, and \emph{Stairs}, where baselines suffer the most severe forgetting. On \emph{Office}, our method reaches 81.8\% while the next best (iCaRL at 75.5\%) lags by over 6\%. On \emph{Stairs}, our method achieves 56.2\%, nearly doubling the performance of Buff-CS (31.3\%) and GDR (32.7\%). These scenes share characteristics that make replay quality critical: strong viewpoint similarity, repetitive textures, and complex spatial layouts that cause naive buffers to over-represent a narrow set of viewpoints. On the easier scenes, such as \emph{Chess}, \emph{Fire}, and \emph{Heads}, our method remains competitive with the baselines, confirming that SARS and SPDD do not sacrifice performance on well-covered scenes. 

\noindent \textbf{Results on i12S.}
The advantage becomes even more pronounced on the i12S dataset as shown in Table~\ref{tab:i12s}. Our method reaches 91.8\% final accuracy, compared with 83.1\% for Buff-CS, 78.7\% for GDR, 66.0\% for iCaRL, and 13.5\% for GEC. The corresponding TFR is 8.1, again the lowest among all continual-learning methods. The gap to Joint narrows to only 5.1 points, which is notable given that i12S contains twelve scenes and much larger spatial extent than i7S. The result suggests that the proposed framework scales well to longer scene sequences and larger spatial coverage.

As shown in Fig.~\ref{fig:perscene}~(middle), the pattern is consistent but more pronounced due to the larger scale. Our method dominates on the most forgetting-prone scenes: \emph{Bed}, \emph{Luke}, \emph{Gate381}, \emph{Lounge}, \emph{Manolis}, and \emph{Floor.5a} all show substantial improvements over baselines. On \emph{Luke}, our method achieves 86.5\% while Buff-CS reaches only 67.3\% and iCaRL drops to 42.6\%. GEC again fails catastrophically on most scenes, confirming that classification-style regularization does not transfer to pose regression at this scale. Notably, our method maintains near-perfect accuracy on scenes where all baselines also perform well, such as \emph{Kitchen-2}, \emph{Living-2}, \emph{Floor.5b}, demonstrating that our gains on difficult scenes do not come at the cost of easy ones.
\begin{figure}[htp]
    \centering
    \includegraphics[width=0.48\textwidth]{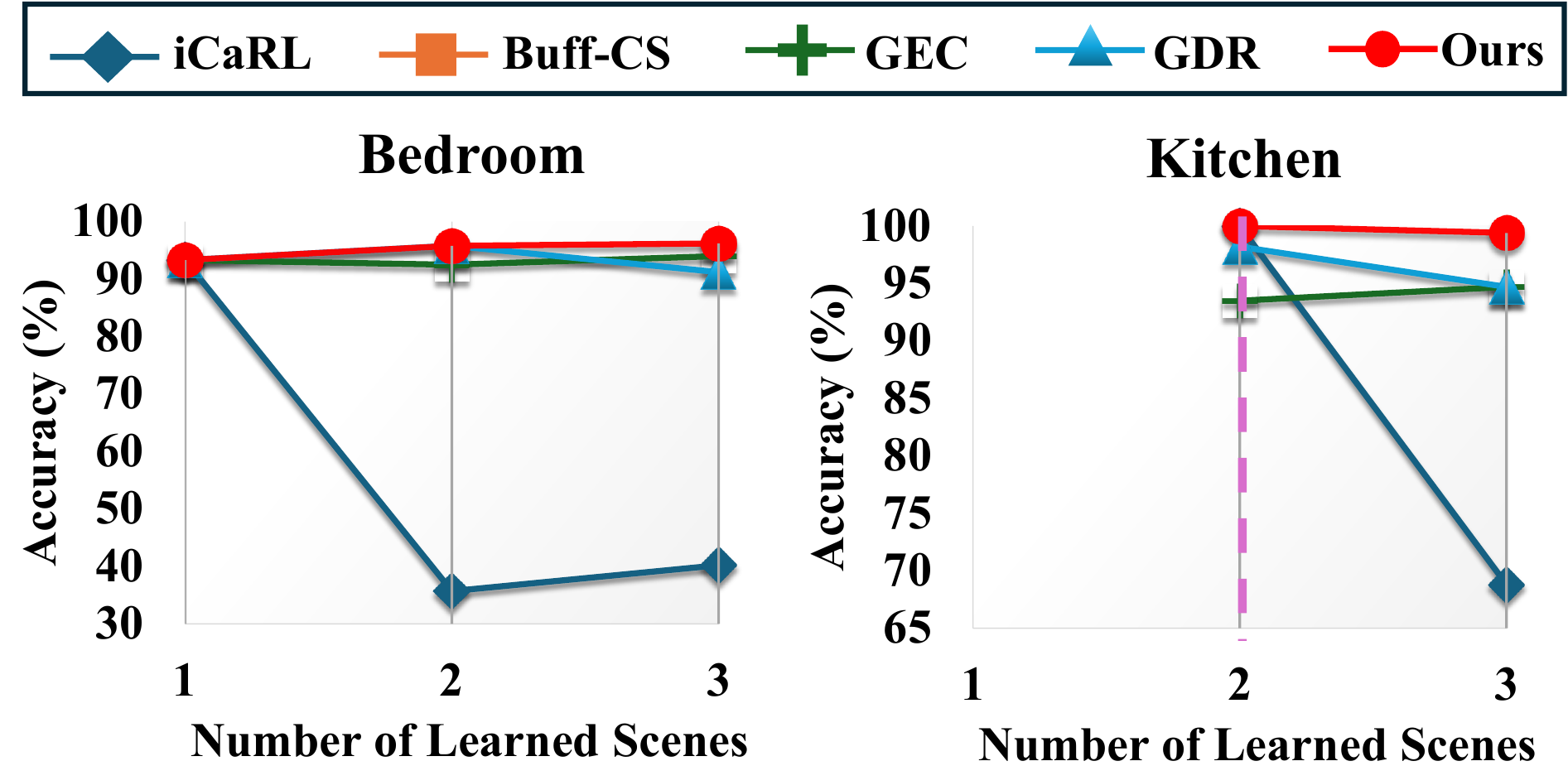}
    \caption{Task forgetting rate comparisons on the Simulation dataset.}
    \label{fig:simtfr}
\end{figure}
\begin{figure*}[htp]
    \centering
    \includegraphics[width=\textwidth]{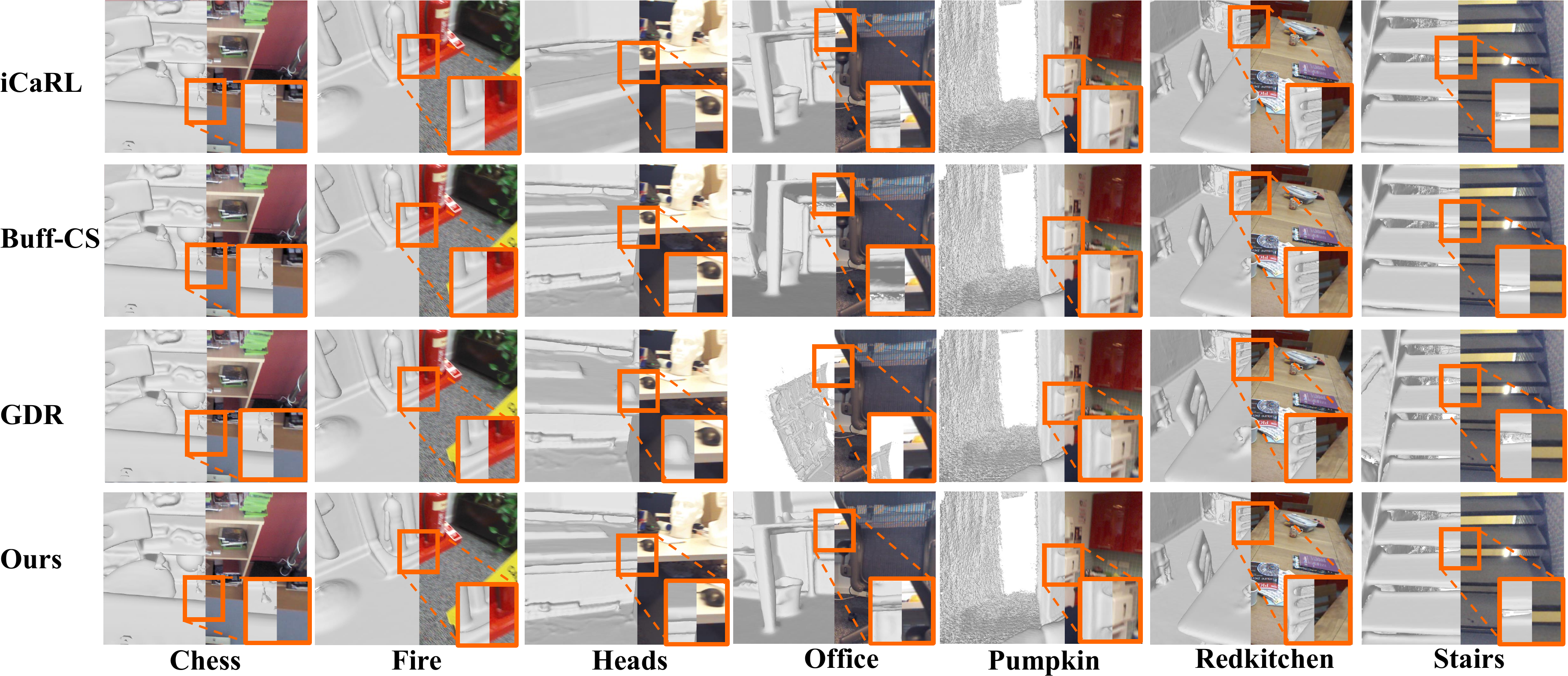}
    \caption{Qualitative localization results on the i7S dataset. For each scene, the left image shows the 3D model rendered using the estimated camera pose (grayscale), and the right image shows the corresponding query image. Accurate alignment between rendered contours and real scene structures indicates precise pose estimation.}
    \label{fig:7v}
\end{figure*}
\begin{figure*}[ht]
    \centering
    \includegraphics[width=1\textwidth]{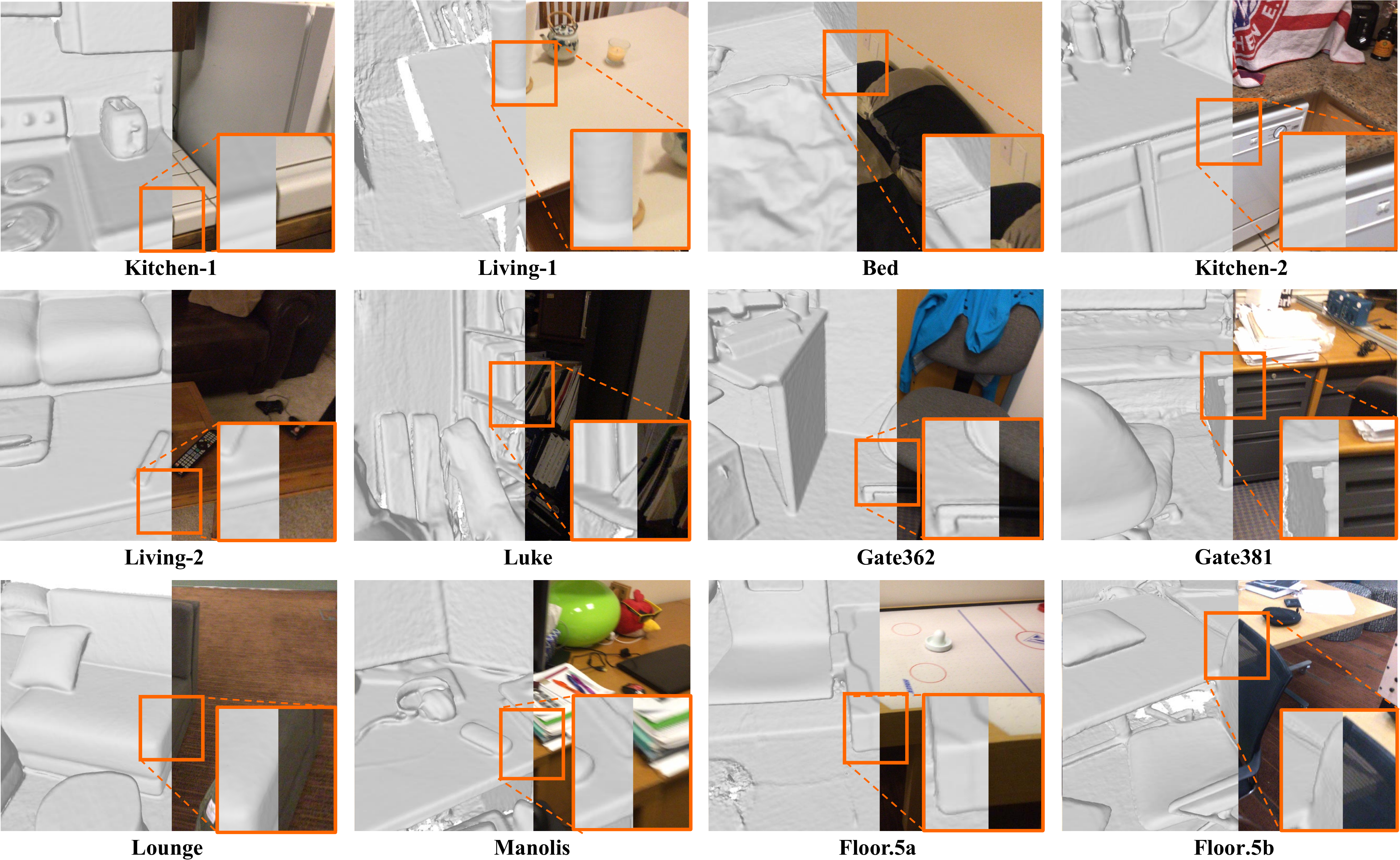}
    \caption{Qualitative localization results on the i12S dataset. Each pair shows the rendered 3D model (left, grayscale) overlaid with the query image (right) using the estimated pose after continual learning across all 12 scenes.}
    \label{fig:12v}
\end{figure*}
\begin{figure}[htp]
    \centering
    \includegraphics[width=0.46\textwidth]{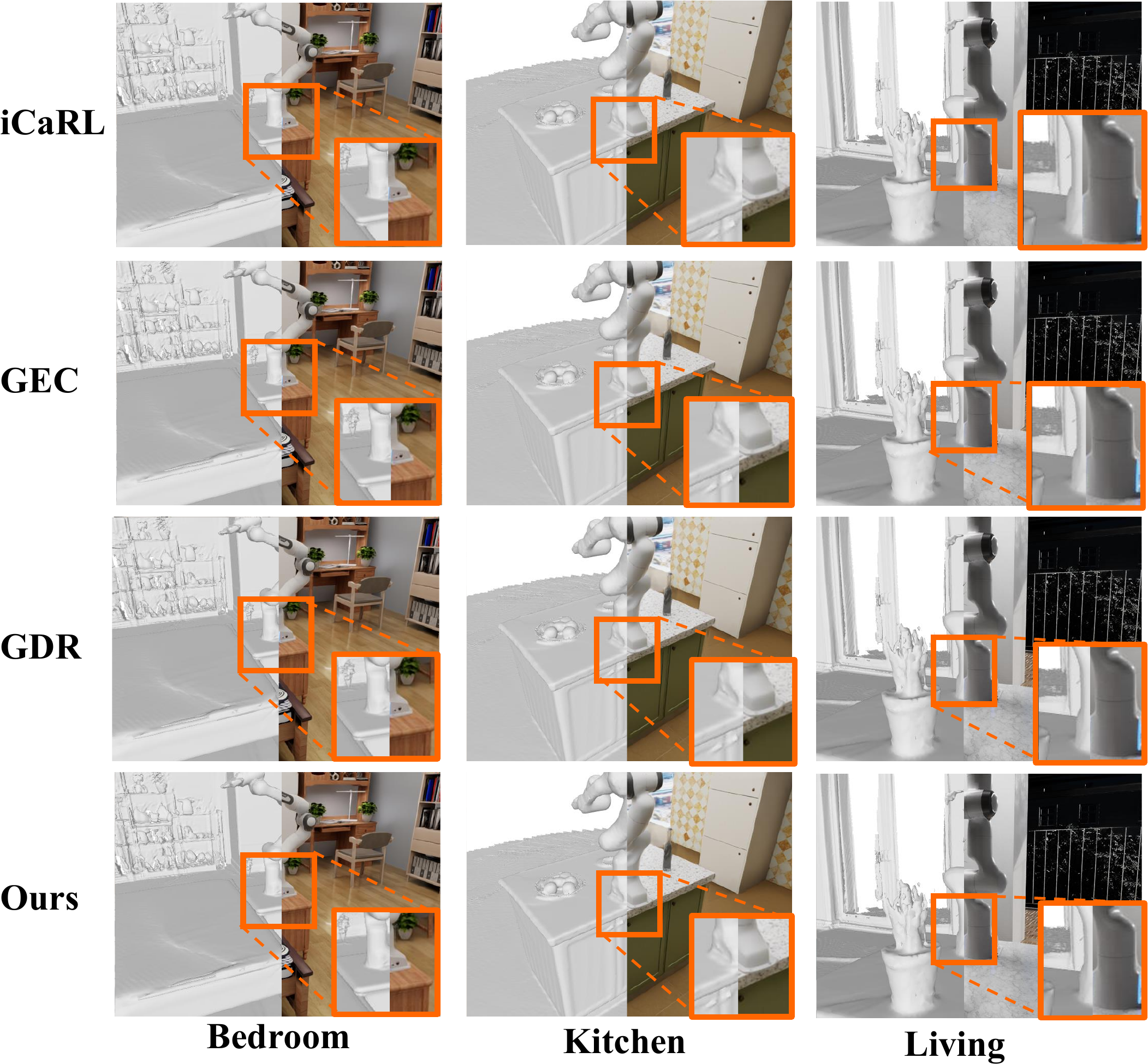}
    \caption{Visualization of localization results on the Simulation dataset. The query image (right) with a rendered image (left, grayscale) using the estimated pose and the ground truth 3D model. }
    \label{fig:simv}
\end{figure}

\noindent \textbf{Results on Sim.}
On the simulated robotic manipulation dataset (Table~\ref{tab:sim}), our method achieves 98.4\% average accuracy with a TFR of only 1.6\% under the stricter $1$ cm/$1^\circ$ criterion, closely approaching the joint-training upper bound of 99.9\%. Compared to GEC (95.0\%) and GDR (95.3\%), which are the competitive baselines on this dataset, our method provides consistent improvements across all three scenes while achieving substantially lower forgetting. This result validates that our framework generalizes effectively to the hand-eye calibration domain with robotic manipulation scenarios.

As shown in Fig.~\ref{fig:perscene}~(right), our method achieves near-ceiling performance across all three scenes, with the smallest variance between scenes. iCaRL shows a dramatic accuracy drop from Living (99.6\%) to Bedroom (40.2\%), revealing that its herding-based selection fails to construct geometrically representative buffers for the robotic manipulation setting. Our method maintains 96.2\% even on the most challenging Bedroom scene, confirming the effectiveness of SARS in the robotic hand-eye calibration domain.

\subsection{Forgetting Rate Analysis}
Fig.~\ref{fig:i12stfr},~\ref{fig:i7stfr}, and~\ref{fig:simtfr} visualize the evolution of per-scene accuracy as the number of learned scenes increases, thereby visualizing forgetting over the whole sequence. 
The final scene in each benchmark is omitted because it has no subsequent tasks and therefore no meaningful forgetting trajectory.

Across all three datasets, the curves of the proposed method degrade more slowly and remain flatter after new scenes are introduced. On the i7S dataset, the advantage is especially clear on \emph{Heads}, \emph{Office}, and \emph{Redkitchen}, where several baselines exhibit sharp accuracy drops after later scenes are learned. On the i12S dataset, the benefit becomes even more pronounced on scenes such as \emph{Living-1}, \emph{Luke}, \emph{Lounge}, \emph{Manolis}, and \emph{Floor.5a}, for which the proposed method maintains substantially higher accuracy throughout the remainder of the sequence. On the Sim dataset, the trajectories are nearly flat after each scene update, consistent with the low final TFR of 1.6\%. Together with the TFR summaries in Tables \ref{tab:i7s},~\ref{tab:i12s}, and~\ref{tab:sim}, these plots show that the proposed framework reduces both immediate post-update forgetting and cumulative long-term drift.

\subsection{Qualitative Visualization}
Fig~\ref{fig:7v},~\ref{fig:12v}, and~\ref{fig:simv} provide qualitative evidence of localization accuracy by overlaying 3D model renderings with query images. For each scene, the left image shows the ground-truth 3D model rendered from the estimated camera pose (grayscale), and the right image shows the corresponding real query image. When the estimated pose is accurate, the rendered contours align precisely with the real scene structures.

On the i7S dataset, as shown in Fig.~\ref{fig:7v}, the renderings show alignment across all seven scenes. 
On the easier scenes such as \emph{Chess}, \emph{Fire}, and \emph{Heads}, the contours are nearly indistinguishable from the ground truth, confirming that our method achieves near-joint performance on these scenes. On the more difficult scenes such as \emph{Office}, \emph{Redkitchen}, and \emph{Stairs}, although the result of our method shows alignment error, the camera pose median error of our method is lower compared to the continual learning baselines.

On the i12S dataset, Fig.~\ref{fig:12v} shows that our method is able to maintain consistent alignment across 12 diverse indoor environments. Even in scenes with relatively complex spatial structures, such as \emph{Luke}, \emph{Gate362}, and \emph{Lounge}, the rendered contours largely agree with the real scene geometry, indicating that the framework can scale to longer continual learning sequences without obvious degradation in pose quality

On the Sim dataset, Fig.~\ref{fig:simv} provides qualitative evidence for the hand-eye calibration objective. In the \emph{Bedroom}, \emph{Kitchen}, and \emph{Living} environments, the rendered robotic arms and scene objects remain broadly aligned with the ground-truth images, suggesting that the estimated camera-to-robot transformations are reasonably preserved after sequential scene updates.

\subsection{Ablation Study}
\begin{table}[]
    \centering
    \caption{Ablation study on i7S and i12S. Each row adds one module to the iCaRL baseline.}
    \label{tab:ablation}
\begin{tabular}{@{}lll@{}}
\toprule
Methods\textbackslash{}Datasets & i7S     & i12S    \\ \midrule
iCaRL (baseline)  & 65.2 & 66.0 \\
+ SARS & 72.0\red{6.8} & 90.9\red{24.9} \\
+ SARS + SPDD & 74.9\red{2.9} & 91.8\red{0.9} \\ \bottomrule
\end{tabular}
\end{table}

Table~\ref{tab:ablation} analyzes the contribution of the two proposed components by incrementally adding them to the replay-based baseline used in the ablation protocol.

\noindent \textbf{Effect of SARS.}
Adding SARS to the baseline improves average accuracy from 65.2\% to 72.0\% (+6.8\%) on i7S and from 66.0\% to 90.9\% (+24.9\%) on i12S. The dramatic gain on i12S, which contains more scenes and a larger spatial extent, confirms that geometric coverage of camera poses is far more critical than sample randomness for anti-forgetting in localization.

\noindent \textbf{Effect of SPDD.}
Adding SPDD on top of SARS further improves performance to 74.9\% on i7S and 91.8\% on i12S, corresponding to gains of +2.9\% and +0.9\% over the SARS-only variant. Although the incremental gain is smaller than that of SARS, it is consistent across both datasets. This suggests that once the replay buffer becomes more representative, structured distillation could alleviates the pose precision degradation by preserving the internal geometry of localization predictions during rehearsal.

A useful observation from Table~\ref{tab:ablation} is that the dominant gain comes from the spatial-aware replay design, while the structured distillation term provides a further refinement. This is consistent with the broader empirical trend of the paper: continual hand-eye calibration benefits most when the replay mechanism respects the structure of pose space, and when the supervision applied to replayed samples preserves more than a single pose output.


\section{Conclusion}
\label{sec:conclusion}
We present a continual hand-eye calibration framework for open-world robotic manipulation. By combining geometry-aware replay with structure-preserving distillation, the proposed method enables a single visual localization model to adapt to new scenes while retaining accurate pose estimation on previously learned environments. Across two public continual-localization benchmarks and a simulated manipulation dataset, the method consistently outperformed strong continual-learning baselines and reduced forgetting throughout the sequential learning process. These results suggest that effective continual calibration requires both representative pose-space memory and preservation of structured localization knowledge during replay.
A current limitation is that the evaluation is mainly restricted to indoor scenes and simulation. Extending the framework to real-robot deployment and larger-scale, less structured environments is an important direction for future work.



%

\bibliography{IEEEabrv,references} 
\bibliographystyle{IEEEtran}

\vfill

\end{document}